%% file: arxiv-version.tex
\documentclass[sigconf]{acmart}

\AtBeginDocument{%
  \providecommand\BibTeX{{%
    \normalfont B\kern-0.5em{\scshape i\kern-0.25em b}\kern-0.8em\TeX}}}

\usepackage{multirow}
\usepackage{makecell}
\begin{document}

\title{Full-Stage Pseudo Label Quality Enhancement for Weakly-supervised Temporal Action Localization}

\author{Qianhan Feng$\dagger$}
\affiliation{%
  \institution{State Key Lab of GAI, SIST, Peking University}
  \city{Beijing}
  \country{China}}
\email{fengqianhan@stu.pku.edu.cn}

\author{Wenshuo Li}
\affiliation{%
  \institution{Huawei Noah's Ark Lab}
  \city{Beijing}
  \country{China}}
\email{liwenshuo@huawei.com}

\author{Tong Lin*}
\affiliation{%
  \institution{State Key Lab of GAI, SIST, Peking University}
  \city{Beijing}
  \country{China}}
\email{lintong@pku.edu.cn}

\author{Xinghao Chen*}
\affiliation{%
  \institution{Huawei Noah's Ark Lab}
  \city{Beijing}
  \country{China}}
\email{xinghao.chen@huawei.com}

\renewcommand{\shortauthors}{Feng and Li, et al.}

\begin{abstract}
  Weakly-supervised Temporal Action Localization (WSTAL) aims to localize actions in untrimmed videos using only video-level supervision. Latest methods introduce pseudo label learning framework to bridge the gap between classification-based training and inferencing targets at localization, and achieve cutting-edge results. In these frameworks, a classification-based model is used to generate pseudo labels for a regression-based student model to learn from. However, the quality of pseudo labels, which is a key factor to the final result, is not carefully studied. In this paper, we propose a set of simple yet efficient pseudo label quality enhancement mechanisms to build our \textbf{FuSTAL} framework. FuSTAL enhances pseudo label quality at three stages: cross-video contrastive learning at proposal \textit{Generation-Stage}, prior-based filtering at proposal \textit{Selection-Stage} and EMA-based distillation at \textit{Training-Stage}. These designs enhance pseudo label quality at different stages in the framework, and help produce more informative, less false and smoother action proposals. With the help of these comprehensive designs at all stages, FuSTAL achieves an average mAP of 50.8\% on THUMOS'14, outperforming the previous best method by 1.2\%, and becomes the first method to reach the milestone of 50\%. Code is available at https://github.com/fqhank/FuSTAL.
\end{abstract}

  


\maketitle

\section{Introduction}
Temporal Action Localization (TAL) \cite{DBLP:conf/iccv/XuDS17,DBLP:conf/iccv/SridharQMLDL21,DBLP:conf/cvpr/XuZRTG20,DBLP:conf/eccv/ZhangWL22,DBLP:conf/cvpr/ShiZCMLT23} aims at predicting the start and end of the actions within an arbitrarily long untrimmed video, and classify the snippets simultaneously. The application scenarios of TAL are very wide, ranging from robotic, industrial safety inspection, sports to video summarization \cite{DBLP:conf/cvpr/LeeGG12,DBLP:journals/vc/VishwakarmaA13}. While it is challenging for the model to obtain a strong temporal and semantic understanding, the reliance on the dense temporal annotation is even more expensive. Manual annotation including start and end points of each action within every video are required, which makes labeling labor-consuming. Besides, non-professional artificial labeling could largely damage the learning.

\begin{figure}[t]
  \centering
  \includegraphics[width=1.0\columnwidth]{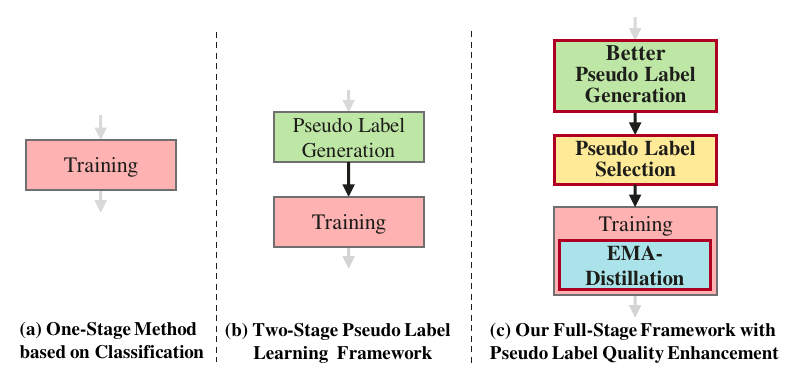}
  \vspace{-20pt}
  \caption{Unlike classical two-stage pseudo label learning framework for Weakly-supervised Temporal Action Localization (WSTAL), we excavate potential stages in the framework and enhance pseudo label quality at full-stage.}
  \vspace{-15pt}
  \label{fig:headfig}
\end{figure}

To mitigate the need for dense annotations, Weakly-supervised Temporal Action Localization (WSTAL) is proposed. With no temporal action localization information available, WSTAL methods apply various tricks to conduct feature extraction and learning, thus aligning with the video-level category labels. While some WSTAL algorithms adapt attention mechanism \cite{DBLP:conf/cvpr/ShiDMW20,DBLP:conf/aaai/XuZCXNPW19}, others perform weakly-supervised learning by developing a Multiple Instance Learning (MIL) scheme \cite{DBLP:conf/cvpr/WangXLG17,DBLP:conf/aaai/LeeUB20,DBLP:conf/cvpr/MaGVY21,DBLP:conf/mm/MoniruzzamanYHQ20,zhang2021cola,Ren_2023_CVPR} with various information mining tricks including contrastive learning and prior constraint. However, these one-stage strategies are classification-based, producing final predictions by a series of post-processing on small snippets, including thresholding, merging and NMS. This artificial post-processing design is far different from directly localizing actions, which is the final goal of our task. The gap between classification and localization makes these algorithms lack knowledge of the action as a complete whole, but focusing more on micro features on each small time-step. To bridge this gap, some latest approaches \cite{DBLP:conf/cvpr/RizveM0HSSC23,DBLP:conf/cvpr/ZhouHWLL23} design two-stage schemes to generate pseudo labels to train a regression-based student model. The introducing of pseudo label learning framework brings significant performance boosting. However, the quality of pseudo labels, which is the most important key factor in the framework, is seldom specifically studied. Previous works mostly pay attention to only proposal generation, but neglect the potential of creating better action proposals in other stages as well as the alignment between pseudo label information and localization goal. Even at pseudo label generation stage, the generated action proposals are not informative enough because of limiting attention within a single video example. 

To exploit the potential of pseudo labels in WSTAL pseudo label learning framework, we view the framework as a multi-stage process: including \textit{Generation-Stage} for action proposal generation, \textit{Selection-Stage} for filtering out noisy false positive action proposals and \textit{Training-Stage} for regression-based student model training and promotion. We suggest that promoting action proposals only in \textit{Generation-Stage} is not enough. Therefore, we propose to enhance pseudo labels, or namely the action proposals, in all three stages.  

In \textit{Generation-Stage} for proposal generation, we introduce cross-video information to help find more essential characteristics of actions. Previous works \cite{DBLP:conf/cvpr/RizveM0HSSC23,DBLP:conf/cvpr/ZhouHWLL23,zhang2021cola,Ren_2023_CVPR} are limited to mining information from individual videos. However, with only in-video snippets available for contrast, some confusing segments might be misclassified, since the scenes and characteristics are relatively similar within one video. With the aid of other videos, it would be easier to tell the target action from the background. Although some previous methods do utilize cross-video information, they are not suitable to be the proposal generator in a pseudo label learning framework. RSKP \cite{DBLP:conf/cvpr/Huang0022} uses intra-video representations to update original feature, only to refine the classification. DCC \cite{DBLP:conf/cvpr/LiYJW022} utilizes inefficient contrastive learning but neglects boundary information, which is the key information that student model needs. We introduce a straight forward and efficient approach to utilize cross-video similarities to help generate action snippet proposals with better quality. To be more specific, we select video pairs with the same video-level label, and mine the embeddings of easy and hard actions and backgrounds based on MIL training. Then, a contrastive loss is applied to let the mined hard embeddings of one video approach easy snippet features from another video in the pair.  

After \textit{Generation-Stage}, previous frameworks would directly start \textit{Training-Stage}, feeding all generated proposals to student model. Even some take proposal quality into consideration, only easy confidence thresholding is used. However, we suggest that there are still huge amount of false positives in proposals that actually do not overlap with any ground-truth actions, which could mislead action-level information learning and damage the performance. This is partly because many backgrounds tend to appear along with the action, thus resulting in similar high activation. Nevertheless, there is no reasonable method to prevent these harmful proposals entering pseudo label training in existing works. To fulfill this absence, we add a specially designed \textit{Selection-Stage} before \textit{Training-Stage} for pseudo label filtering. We analyze the distribution of these false positives and introduce a simple yet efficient prior based mechanism to filter out as many false positives as we could. We find that the generated predictions around backgrounds tend to be less dense than those highly overlapped with ground-truth actions. Based on this prior, we calculate the IoU matrix of all proposals within a video, and filter out all proposals with low proposed IoU scores. Only proposals with IoU score higher than threshold are qualified to become pseudo labels for student model to learn from.  

At \textit{Training-Stage}, a regression-based student model is trained in a supervised manner, and previous works stop when the training ends. However, we believe the latent of student itself is not totally excavated. The original pseudo labels come from classification-based models, who do not directly give out the start and end time point of actions. Instead, they classify on each small snippets and rely on artificial post-processing to generate localization prediction, and do not have good understanding of action as a whole. On the contrary, the student model is regression-based, and is directly trained with macro information of actions. Even not accurate enough, student model views actions as a whole from the beginning, thus maintaining a better understanding of the complete action than original pseudo label generator. Therefore, after the training of original pseudo label reaches its ceiling and no further improvement could be seen, we turn to the Exponential Moving Average (EMA) of the regression-based student for more precise pseudo label generation. With smoother and more accurate new action proposals, a fast final pseudo label training is conducted. 

In test time, only the trained regression-based student model is used for inference. We conduct extensive experiments on standard WSTAL datasets, THUMOS'14, achieving 1.2\% improvement on the average mAP over the previous state-of-the-art method. What's more, our method is the first one to reach the milestone of 50\% in WSTAL task with only video-level category label accessible. To summarize, our work makes the following contributions: 
\begin{itemize}
    \item We provide a new insight about potential stages for proposal enhancement in WSTAL pseudo label learning framework. Furthermore, we propose novel strategies in each stage to produce better action proposals.
    \item An easy yet efficient cross-video contrastive learning mechanism is proposed to produce more informative original pseudo labels. To excavate the latent of student model itself, an EMA-distillation is applied to generate smoother pseudo labels at late training. 
    \item We design a prior-based proposal filtering mechanism to filter out false positives, fulfill the absence of reliable pseudo label selection method in WSTAL frameworks.  
    \item Our method outperforms all previous WSTAL methods on THUMOS'14 dataset, and becomes the first one to reach the milestone of the average mAP of 50\%.
\end{itemize}

\section{Related Work}
\textbf{Fully-supervised Temporal Action Localization.} Temporal Action Localization (TAL) aims to localize and classify all actions from an untrimmed video. Most existing methods can be broadly divided into two categories: two-stage methods and one-stage methods. The two-stage methods \cite{DBLP:conf/cvpr/QingSGW0W0YGS21,DBLP:conf/cvpr/ShouWC16,DBLP:conf/iccv/XuDS17,DBLP:conf/iccv/ZhuT00021,DBLP:journals/ijcv/ZhaoXWWTL20,DBLP:conf/iccv/SridharQMLDL21,DBLP:conf/cvpr/XuZRTG20} comprise proposal generation stage and classification stage. This kind of methods enhance their performance by improving either the quality of proposals \cite{DBLP:conf/iccv/ZhuT00021,DBLP:conf/iccv/ZengHGTRZH19,DBLP:conf/cvpr/ShouWC16} or the robustness of classifiers \cite{DBLP:conf/iccv/XuDS17,DBLP:journals/ijcv/ZhaoXWWTL20}. On the other hand, one-stage methods generate and classify the candidates simultaneously. Some of one-stage methods build this hierarchical architecture with the convolutional network (CNN) \cite{DBLP:conf/cvpr/Lin0LWTWLHF21,DBLP:journals/tip/YangPZFH20,DBLP:journals/cviu/YangCZLW23}, while some others have achieved remarkable performance by introducing Transformer architecture \cite{DBLP:journals/pami/GaidonHS13,DBLP:conf/eccv/WengPHCZ22,DBLP:conf/eccv/ZhangWL22,DBLP:conf/cvpr/ShiZCMLT23}.  

\textbf{Weakly-supervised Temporal Action Localization.} Despite the success of Fully-supervised TAL methods, they require massive and consuming frame-level artificial annotations, which limits their ability to transfer among real-world data. Weakly-supervised Temporal Action Localization is then proposed to solve this problem, which only requires video-level category labels. With only video-level labels available, UntrimmedNet \cite{DBLP:conf/cvpr/WangXLG17} firstly introduces a Multiple Instance Learning (MIL) framework to handle the problem by classifying top activated snippets to generate proposals. From then on, a number of MIL-based methods \cite{DBLP:conf/aaai/LeeUB20,DBLP:conf/cvpr/MaGVY21,DBLP:conf/mm/MoniruzzamanYHQ20,Ren_2023_CVPR} are proposed to enhance the quality of classification-based proposals. For example, W-TALC \cite{DBLP:conf/eccv/PaulRR18} learns compact intra-class feature representations by pulling features of the same category to be closer while pushing those of different categories away. CoLA \cite{zhang2021cola} excavates easy and hard snippets within the same video and applies contrastive learning on the features to refine the representation. DCC \cite{DBLP:conf/cvpr/LiYJW022} applies cross-video contrastive learning to WSTAL, but it focuses on contrasting between categories, neglecting the boundary information, and requires huge computation resources to conduct the algorithm. PivoTAL \cite{DBLP:conf/cvpr/RizveM0HSSC23} builds a pseudo label learning framework, and introduces a series of prior knowledge like the correlation between background scene and the action to generate better pseudo action snippets, then use them to train a regression-based head directly. Although PivoTAL achieves edge-cutting results, the pseudo labels in the framework are used flatly. The lack of study on better pseudo label quality leaves space for further improvement.  

\textbf{Pseudo Label Learning.} Pseudo label learning strategy is widely used in weakly-supervised problems. When the example is not annotated, producing a pseudo label through confidence score thresholding is a commonly used methodology \cite{lee2013pseudo,DBLP:conf/nips/SohnBCZZRCKL20,DBLP:conf/nips/ZhangWHWWOS21}. Under WSTAL setting, works with pseudo label training mainly focus on pseudo label generation. Most of these methods \cite{DBLP:conf/eccv/Zhai0TZY020,DBLP:conf/aaai/YunQWM24,luo2020weakly,DBLP:conf/cvpr/HeYKCZS22,DBLP:conf/cvpr/LiYJW022} focus on snippet features and generate pseudo labels on snippet-level. Zhou et al. \cite{DBLP:conf/cvpr/ZhouHWLL23} try to include action boundary information in pseudo labels, and design self-correction mechanism to cut down the confidence bias, thus producing pseudo labels with higher quality on action level. PivoTAL \cite{DBLP:conf/cvpr/RizveM0HSSC23} includes prior knowledge to generate informative proposals for regression model to learn. How to select pseudo labels after generation under video studying and the potential improvement during training still remain to be explored.

\section{Proposed Method: FuSTAL}
\subsection{Preliminaries} 
In WSTAL, we only have access to a set of videos with the video-level label available $\mathbb{V}=\{\mathbf{v}^{(i)},\mathbf{y}^{(i)}\}^{N}_{i=1}$, where N is the total number of videos, $\mathbf{v}^{(i)}$ and $\mathbf{y}^{(i)}$ represent untrimmed video and action category label. There is no information about the precise start and end time points of actions, and it is also unknown that how many actions are there in one video. $\mathbf{y}^{(i)}$ is represented as a multi-hot label encoding $\mathbf{y}^{(i)}\in\{0,1\}^{C}$, where $C$ is the number of action categories. The target in inference time is to generate a set of predictions of action snippets $\mathbb{A}^{(i)}=\{s_{j},e_{j},c_{j}\}^{M}_{j=1}$ for video $\mathbf{v}^{(i)}$, $s_{j}$ and $e_{j}$ are the start and end time of action $\mathbf{a}_{i,j}$, while $c_{j}$ is the predicted category.  

Given an input untrimmed video $V_{n}$, we follow the common process \cite{DBLP:conf/aaai/LeeUB20,zhang2021cola} to divide it into multiple snippets, \textit{i.e.}, $V_{n}=\{S_{n,l}\}^{L_{n}}_{l=1}$. During training, we sample a fixed number of $T$ snippets $\{S_{n,l}\}^{T}_{l=1}$ from the variational length. After sampling, a pre-trained feature extractor (\textit{e.g.}, I3D \cite{DBLP:conf/cvpr/CarreiraZ17}) is applied to extract the RGB features $X^{R}_{n}=\{x^{R}_{t}\in\mathbb{R}^{d} \}^{T}_{t=1}$ and optical features $X^{O}_{n}=\{x^{O}_{t}\in\mathbb{R}^{d} \}^{T}_{t=1}$ from the snippets set respectively, and $d$ is the feature dimension of each snippet. The extracted features of two modals are then concatenated to generate the input feature $X_{n}\in\mathbb{R}^{T\times2d}$ for the network.  

\subsection{Insight and Overview}
Without annotation for each action, one classical method is to conduct classifying on each small snippet along temporal dimension. However, this classification-based route relies on a lot of artificial operations like thresholding and NMS to generate final action proposals from snippet-level classifying predictions. These operations progressively discard the knowledge of learned action characteristics. To bridge the gap between classification operation and ultimate localization goal, pseudo label learning framework for WSTAL is designed. In the framework there is a classification-based action proposal generator and a regression-based student model who directly learns from the pseudo labels. Different from classification-based methods which focus on tiny snippets mergence, regression-based models view the action as a whole from a macro perspective and propose regression predictions directly, achieving better maintenance of understanding of actions as a whole.

However, pervious works \cite{DBLP:conf/cvpr/RizveM0HSSC23} simply treat the framework as a two-stage algorithm, and focus mainly on action proposal generation. We suggest that the pseudo label quality in the framework, which is the most important key factor, is not only related to generation, but also can be promoted from other aspects. Being aware of the latent promotion pseudo label quality left behind, we are motivated to regard the pseudo label learning framework in WSTAL from a new view: instead of two stages, we suggest that the framework can be separated into \textit{Generation-Stage} for action proposal generation, \textit{Selection-Stage} for filtering out noisy proposals and \textit{Training-Stage} for training student as well latent promotion.  

\begin{figure*}[htbp]
  \centering
  \includegraphics[width=2.0\columnwidth]{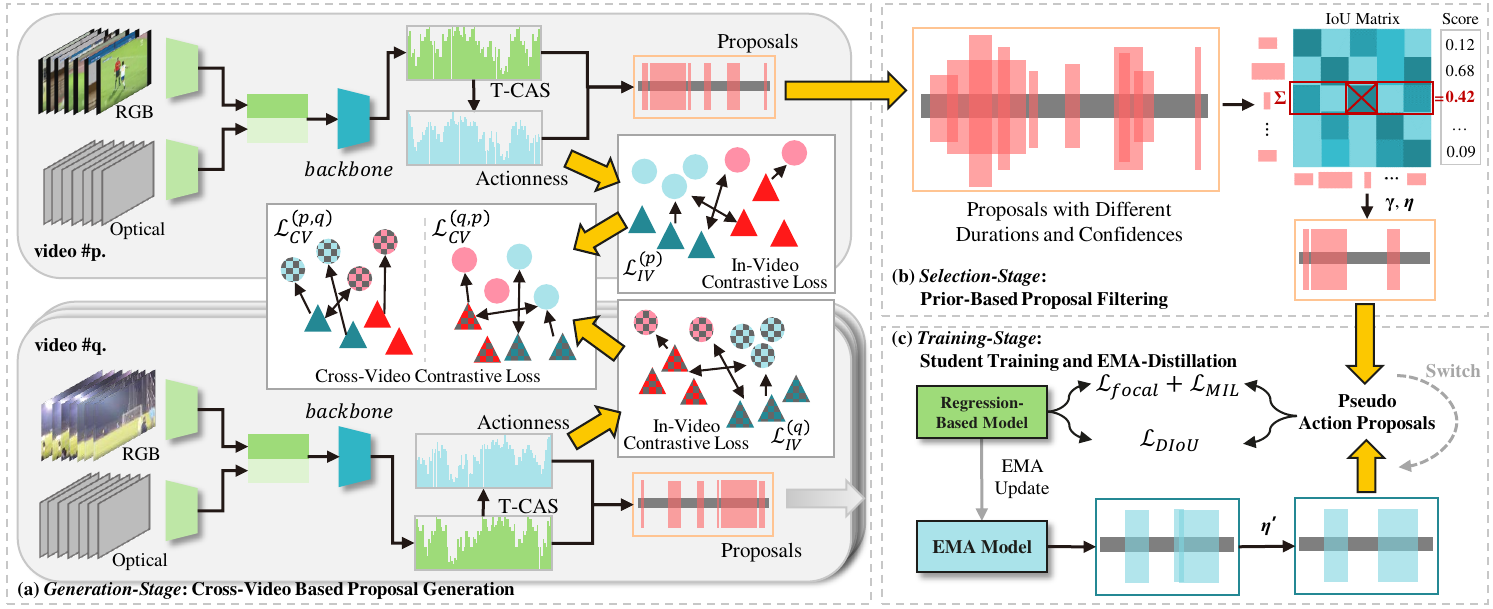}
  \vspace{-10pt}
  \caption{Overview of FuSTAL
  : (a) \textit{Generation-Stage:} An in-video and a cross-video contrastive losses are applied on mined snippets to help excavate the essential characteristics, thus generating more informative action proposals. Triangles and circles in different colors and patterns refer to hard and easy embeddings from different videos. (b) \textit{Selection-Stage:} The initial proposals are gathered to calculate an IoU score for each. Only proposals with score higher than thresholds would be kept as pseudo labels. (c) \textit{Training-Stage:} A regression-based student model is trained with selected action proposals in supervised manner. Meanwhile, an EMA model is updated, and switches to become new label generator once original proposals reach their ceiling. Only the trained regression-based model is used for inference.
  }
  \label{fig:main}
  \vspace{-10pt}
\end{figure*}

With new insight about stages, we present a novel pseudo label learning framework for WSTAL, namely FuSTAL. In FuSTAL, each stage is armed with a special mechanism for pseudo label quality enhancement. The pseudo label is firstly generated by a cross-video-based generator and then filtered by a prior-based mechanism. The selected proposals are given to a student model to learn from. Finally, the trained student model gets smoother pseudo labels from self-distillation for final promotion at late \textit{Training-Stage}. At inference time, only the regression-based student model is used. The overview of FuSTAL is shown in Figure \ref{fig:main}, and we illustrate novel methods for better pseudo labels at \textit{Generation-Stage}, \textit{Selection-Stage} and \textit{Training-Stage} in detail in the following sections.  

\subsection{Cross-Video Based Proposal Generation}
With no accurate action annotation available for regression-based model to learn, we need to train a classification-based model to generate action proposals, or namely pseudo labels. We recognize this stage as \textit{Generation-Stage}.  

Given the cross-modal video feature $X_{n}$ as input, it is a common way to train a network to extract the Temporal Class Activation Sequence (T-CAS) $\mathcal{A}_{n}$, which represents the action classification prediction at each temporal node. Since the category label for each temporal node is not available, we follow prior works \cite{DBLP:conf/eccv/PaulRR18,zhang2021cola,DBLP:conf/cvpr/RizveM0HSSC23} to train the network with video-level label using a Multiple Instance Learning (MIL) strategy. To be more specific, for each class $c$, top-$k$ scores along the temporal dimension are selected out and summed up to compute the confidence score for this class. Then the $\mathrm{SoftMax}$ function is applied on $C$ class scores to get the video-level prediction, which is used to calculate the MIL loss in the cross-entropy form:
\begin{equation}
    \label{MIL}
    \mathcal{L}_{MIL} = -\frac{1}{N}\sum_{n=1}^{N}\sum_{c=1}^{C}\hat{y}_{n;c}\log(p_{n;c}),
\end{equation}   
where $\hat{y}_{n;c}\in\mathbb{R}^{C}$ is the video-level category label. The MIL loss helps find the most discriminative segments through classification. 

\begin{figure}[t]
  \centering
  \includegraphics[width=1.0\columnwidth]{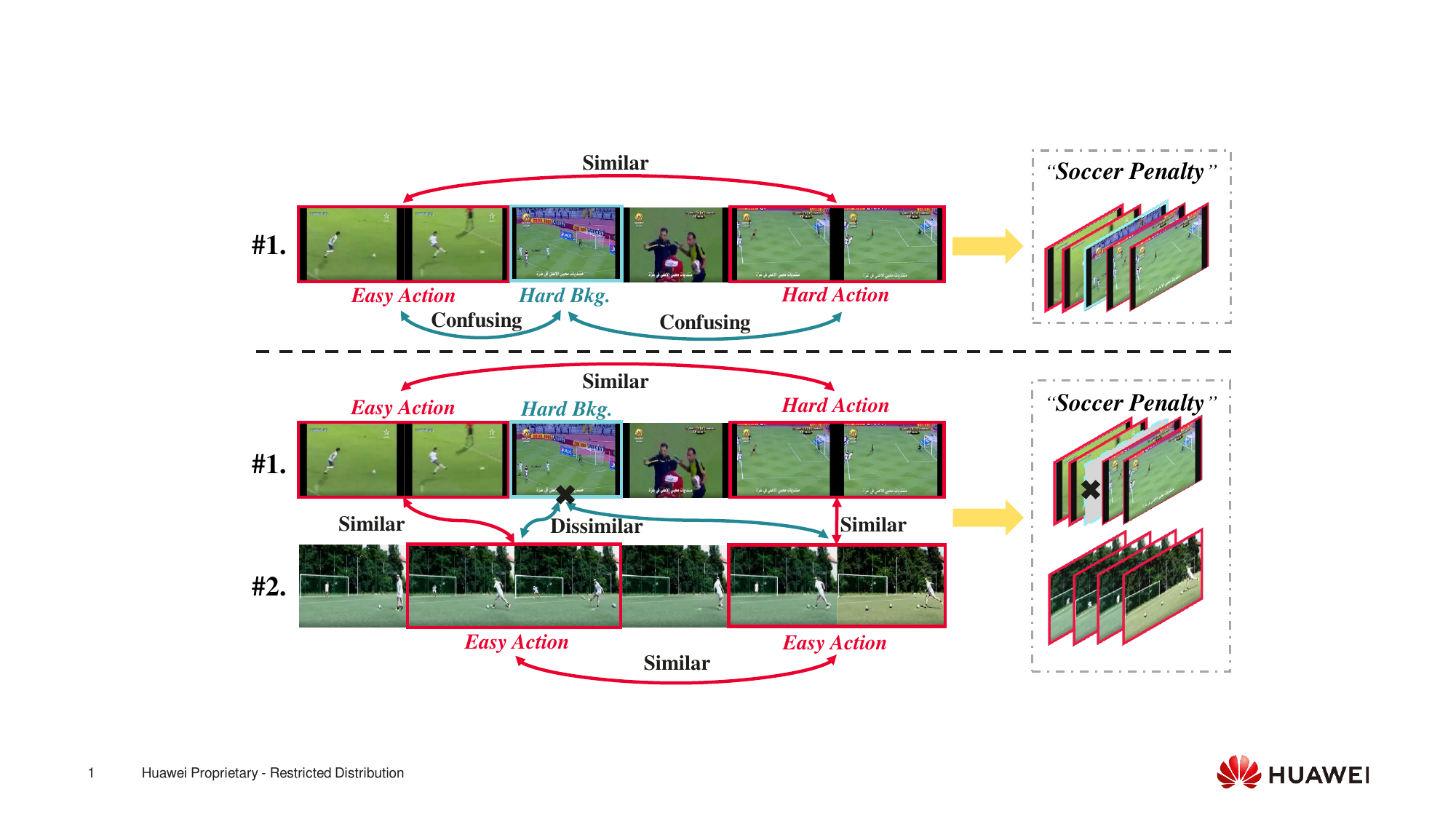}
  \vspace{-20pt}
  \caption{In ``\textit{Soccer Penalty}'' video, background snippets after kicking (blue box) are similar with previous penalty-kicking action. But with another video as reference, this similarity would be highly inhibited, thus excavating the essential characteristics of the action --- run-ups and kicks.}
  \vspace{-15pt}
  \label{fig:fig1}
\end{figure}

Since student model in the framework is regression-based, the time boundary information of action proposal is very important, and should be considered with high priority in pseudo label generation. To create clearer boundaries, network should have great knowledge about the essential characteristics of actions and the difference with background scenes. However, most existing MIL-based methods are restricted within a single video, where limited information could be excavated. What's more, the background scene and the characteristics of actions are relatively stable and similar within a video, which makes it confusing and challenging to notice the most significant difference, just as shown in Figure \ref*{fig:fig1}. 

To tackle this problem, we propose to utilize cross-video information to help excavate useful information and concentrate on more essential characteristics of target action, thus generating more accurate proposals. In our design, the feature of potential action should be close to the action feature not only in the same video, but also close to those from other videos of the same category, and so are the backgrounds. We consider cross-video contrastive learning a suitable strategy to accomplish this goal. Although DCC \cite{DBLP:conf/cvpr/LiYJW022} shares similar idea to conduct cross-video contrastive learning, it is not suitable to be pseudo label generator in the framework. In DCC, positive and negative sample sets are built up by simply average pooling the whole video feature evenly with a pre-defined division. The area where action feature and background feature come from is totally overlapped, and there is no consideration about the key boundary information. What's more, the mechanism in DCC needs a huge memory bank to store every features, and the model has to be trained twice to avoid model collapse. All of these shortages make DCC not qualified to be the pseudo label generator.  

We here propose an efficient cross-video contrastive learning method for action proposal generation. Firstly, we calculate the \textit{Actioness} $\mathcal{A}^{aness}_{n}$, which refers to the likelihood of containing an general action, for every temporal snippets, through summing up T-CAS along class-dimension, followed by a $\mathrm{Sigmoid}$ function:  
\begin{equation}
    \label{actionness}
    \mathcal{A}^{aness}_{n} = \mathrm{Sigmoid}(f_{sum}(\mathcal{A}_{n})), \mathcal{A}^{aness}_{n}\in\mathbb{R}^{T}.
\end{equation}    

Confusion often occurs at the boundary between actions and backgrounds, where snippets close to the boundary are more difficult to tell then snippets in the inner of intervals. To obtain clearer boundary information, hard snippet features are supposed to approach easy snippets of the same scene, action or background. We follow CoLA \cite{zhang2021cola} to mine easy and hard snippets for contrastive learning. For hard snippets mining, $\mathcal{A}^{aness}_{n}$ is firstly binarized, followed by two cascaded dilation or erosion operations to expand or narrow the temporal extent of action intervals. The differential areas are defined as target hard action or background regions:
\begin{equation}
    \begin{split}
    \label{hardmining}
    \mathcal{R}^{easy}_{n}=\left(\mathcal{A}^{bin}_{n};m\right)^{-}-\left(\mathcal{A}^{bin}_{n};\mathcal{M}\right)^{-}, \\
    \mathcal{R}^{hard}_{n}=\left(\mathcal{A}^{bin}_{n};\mathcal{M}\right)^{+}-\left(\mathcal{A}^{bin}_{n};m\right)^{+},
    \end{split}
\end{equation} 
where $\left(.;*\right)^{-}$ and $\left(.;*\right)^{+}$ are erosion and dilation with mask $*$ respectively. $\mathcal{M}$ and $m$ are two masks with different sizes. 

Finally, $k^{hard}$ snippets are sampled from $\mathcal{R}^{hard}$ to form hard action and background set $S^{HA}_{n}$, $S^{HB}_{n}$. For easy snippets, top and bottom $k$ snippets according to descending $\mathcal{A}^{aness}_{n}$ that not belong to $S^{HA}_{n}$ and $S^{HB}_{n}$ are mined to construct easy action and background set $S^{EA}_{n}$ and $S^{EB}_{n}$. An In-Video contrastive Loss is applied to make features of hard snippets approach easy snippets in the same video:
\begin{equation}
    \begin{split}
    \label{In-Video}
    \mathcal{L}_{IV}=&\mathbb{E}_{x\sim S^{HA}_{i},x^{+}\sim S^{EA}_{i},x^{-}\sim S^{EB}_{i}}\ell\left(x,x^{+},x^{-}\right) \\
    +&\mathbb{E}_{x\sim S^{HB}_{i},x^{+}\sim S^{EB}_{i},x^{-}\sim S^{EA}_{i}}\ell\left(x,x^{+},x^{-}\right),
    \end{split}
\end{equation}
$\ell\left(x,x^{+},x^{-}\right)$ is the InfoNCE loss \cite{DBLP:conf/cvpr/He0WXG20} for contrastive learning.  

Except for contrasting within a single video, we expend contrastive learning to cross-video. For video $p$, all videos sharing same action category label from the same mini-batch are selected to form $\left\{q\right\}_{Q}$. Then, a Cross-Video Loss is proposed to make the feature of hard snippets in video $p$ approaches the feature of easy snippet anchors in video $q$ in the same category: 
\begin{equation}
    \begin{split}
    \label{Cross-Video}
    \mathcal{L}^{(p,q)}_{CV}=&\mathbb{E}_{x\sim S^{HA}_{p},x^{+}\sim S^{EA}_{q},x^{-}\sim S^{EB}_{q}}\ell\left(x,x^{+},x^{-}\right) \\
    +&\mathbb{E}_{x\sim S^{HB}_{p},x^{+}\sim S^{EB}_{q},x^{-}\sim S^{EA}_{q}}\ell\left(x,x^{+},x^{-}\right),
    \end{split}
\end{equation}
and vice versa for $\mathcal{L}^{(q,p)}_{CV}$. With $\mathcal{L}_{CV}$, the network could compare confusing snippets with a clear one from another video, thus explicitly getting rid of the interference of similar background scene and overfitted action characteristics. The total loss for video $p$ could then be formulated as: 
\begin{equation}
    \begin{split}
    \label{total}
    \mathcal{L}^{(p)}=\mathcal{L}_{MIL}+\lambda_{1}\mathcal{L}_{IV}+\lambda_{2}\frac{1}{Q}\sum\mathcal{L}_{CV}^{(p,q)},
    \end{split}
\end{equation}
The last term requires hard snippets in video $p$ to approach easy anchors in all videos from $\left\{q\right\}_{Q}$ with same label.

To transfer the classification results along the temporal dimension into action snippets proposals $\mathcal{P}$, a series of post-processing including binarization and Non-Maximum Suppression (NMS) are performed. The mean confidence score $\mathcal{S}^{conf}$ of each proposal is also added into $\mathcal{P}$ for further use.  

\subsection{Prior-Based Proposal Filtering} 
Most previous works would directly start training the regression student model after finishing \textit{Generation-Stage}. However, we suggest that there are still a lot of false positive snippets that actually seldom overlap with any ground-truths. This is because the action snippet proposals obtained through cross-video contrastive learning is still classification-based, and the model tends to focus on the most discriminative parts under the classification pipeline. And these discriminative snippets not only include the most notable actions, but also include some backgrounds that often come along with the actions in videos. For example, in sports videos that include action of \textit{Soccer Penalty}, some of the soccer players often perform a ball picking action before shooting. These related snippets are likely to be misclassified into actions as false positives, which would largely damage the final results. 

It is a common way to filter out proposals with low confidence score, but this is not enough in WSTAL. This is because the high confidence scores often occur in discriminative snippets, including discriminative actions as well as backgrounds, thresholding would still make many false positives left over. The absence of knowledge of video itself makes the selection a dilemma. 

\begin{figure}[t]
  \centering
  \includegraphics[width=1.0\columnwidth]{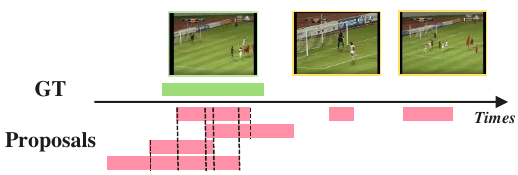}
  \vspace{-15pt}
  \caption{The proposals around ground-truth action segments tend to be more dense than those around backgrounds.}
  \label{fig:dense}
  \vspace{-15pt}
\end{figure}

We inspect the misleading problem of false positives, and insert a novel \textit{Selection-Stage} after \textit{Generation-Stage}. In the new \textit{Selection-Stage}, a prior knowledge of video proposals distribution is incorporated. We notice that although the confidence boundary between actions and backgrounds in proposals is not clear enough, the clustering tendency is relatively discriminative: proposals around true actions tend to be more dense compared with those around backgrounds, which means false positive proposals have lower chances to overlap with others, just as shown in Figure \ref{fig:dense}. That is to say, by selecting proposals with higher overlapping, more false positives would be filtered out. We conduct an experiment to verify this prior knowledge as shown in Figure \ref{fig:distribution}.

To utilize this prior to filter out false positives, we first gather all proposals $\mathcal{P}\in \mathbb{R}^{G}$ from the same video, then calculate the Intersection over Union (IoU) value between each two of them to construct an IoU matrix $\mathbb{M}\in \mathbb{R}^{G\times G}$. Next, we sum up all the IoU values of one proposal to formulate its IoU score:
\begin{equation}
    \label{IoUscore}
    \mathcal{S}^{IoU}_{v}=\sum_{g=1}^{G}\mathbf{1}(g\neq v)\cdot s_{g}, s_{g}\in \mathbb{M}_{v},
\end{equation}
where $\mathbb{M}_{v}$ is the $v$-th row of the matrix, and $\mathbf{1}(g\neq v)$ is 1 when $g\neq v$, otherwise collapses to 0. Although the number of proposals varies from video to video, the distribution of overlapped snippets amount of each proposal is relatively stable, since it has closer relation with local feature rather than global length, and this is why the summarization is used in Equation \ref*{IoUscore} instead of averaging.

Finally, we combine the confidence thresholding and IoU score thresholding to filter out most false positive proposals:
\begin{equation}
    \label{filter}
    \mathcal{P}_{filter} = \left\{p_{v} | p_{v}\in\mathcal{P},\mathcal{S}^{IoU}_{v}\ge\gamma,\mathcal{S}^{conf}_{v}\ge\eta \right\},
\end{equation}
in which $\gamma$ and $\eta$ are thresholds for two scores.  

According to Figure \ref{fig:distribution}, true positive proposals and false positive proposals show reversed distribution tendency in IoU score. However, in some scenarios where actions might be less discernible or more context-dependent, the difference in reversed distribution might be less discriminative. And that is another reason we introduce cross-video contrastive learning in previous \textit{Generation-Stage}: to increase action-background dissimilarity, thus helping cut down the amount of true positives filtered out in \textit{Selection-Stage}. Besides, even some true positive proposals are filtered out under this operation, their information is still well preserved by other overlapped but more complete proposals passing through the filtering. 

After obtaining filtered action proposals with higher proportion of informative true positives, the training of regression-based student model can begin.  

\subsection{Student Model Training and EMA-Distillation for New Proposals}
In our pseudo label learning framework, we aim to train a regression-based student model $\mathbf{D}$, thus bridging the gap between classification and our ultimate objective --- localization. Unlike classification-based model, the regression-based model receives labels including action category and start/end time information. 

We train student model $\mathbf{D}$ in a supervised manner with selected proposals $\mathcal{P}_{filter}$ as hard labels. We incorporate a focal loss $\mathcal{L}_{focal}$ \cite{DBLP:conf/iccv/LinGGHD17} for per-clip action classification. Moreover, to directly find the action boundaries without multi-step processing, a DIoU based regression loss \cite{DBLP:conf/aaai/ZhengWLLYR20} $\mathcal{L}_{DIoU}$ is applied to the regression head of the student model. Finally, an MIL classification loss is adopted to help excavate discriminative information under WSTAL setting. The training loss for regression-based model $\mathbf{D}$ could be formulated as:  
\begin{equation}
    \label{training}
    \mathcal{L}_{Loc}=\mathcal{L}_{focal}+\mathcal{L}_{DIoU}+\mathcal{L}_{MIL}.
\end{equation}

Pseudo labels from $\mathcal{P}_{filter}$ will gradually meet the ceiling of their latent, where student's performance would appear hard to improve anymore. This is because the model has already extracted enough knowledge from quality-limited classification-based labels, and the predictions of the model could surpass far beyond the labels. Traditional pseudo label frameworks would end \textit{Training-Stage} at here, we decide to explore about whether we can create new pseudo labels with better quality to replace original ones. 

Firstly, we revisit previous student training and build an auxiliary network $\mathbf{\hat{D}}$ sharing the same architecture with $\mathbf{D}$. We update its parameters through the Exponential Moving Average (EMA) of $\mathbf{D}$: 
\begin{equation}
  \label{EMA}
  Param_{\mathbf{\hat{D}}}= \alpha\times Param_{\mathbf{\hat{D}}} + (1-\alpha)\times Param_{\mathbf{D}},
\end{equation}
$Param_{\mathbf{\hat{D}}}$ and $Param_{\mathbf{D}}$ are parameters of two models. The EMA model does not feed back to $\mathbf{D}$ when updating the parameters.

Once the training from $\mathcal{P}_{filter}$ stops, we start an auxiliary sub-stage in late \textit{Training-Stage}. In this late sub-stage, we turn $\mathbf{\hat{D}}$ into a teacher to generate new pseudo labels for student model $\mathbf{D}$ to learn. To be more specific, the input video is first fed into $\mathbf{\hat{D}}$ to generate action proposals. Then proposals with confidence lower then $\eta'$ is abandoned, and the rest are provided to $\mathbf{D}$ for learning, using the same loss as Equation \ref*{training}.  

The pseudo labels generated by $\mathbf{\hat{D}}$ are smoother, and more importantly, maintain better information about the complete actions from a macro view. With the help of new pseudo labels, student model enjoys its final promotion. And up to now, we finish the overall FuSTAL framework of pursuing better pseudo labels. 

During the inference time, the proposal generator and the proposal refinement are no longer used, and only the trained regression-based student model $\mathbf{D}$ is maintained. The input video features are directly fed into the regression-based model, and directly output the predicted regression and classification results. 

\section{Experiments}
\subsection{Datasets}
We evaluate our method on two popular action localization benchmark datasets, THUMOS'14 \cite{DBLP:journals/cviu/IdreesZJGLSS17} and ActivityNet v1.3 \cite{DBLP:conf/cvpr/HeilbronEGN15}. Only video-level category labels are accessible when using them. 

\textbf{THUMOS'14} contains untrimmed videos from 20 categories. The length and number of action instances varies greatly. We follow previous works \cite{DBLP:conf/aaai/LeeUB20,zhang2021cola} to use 200 videos from validation set for training and 213 videos from test set for evaluation. 

\textbf{ActivityNet v1.3} is a popular benchmark for TAL with 200 categories. On average, each video contains 1.6 action instances and about 36\% frames are ambiguous action contexts or non-action backgrounds. We follow \cite{DBLP:journals/corr/abs-2104-02967} to use 10024 untrimmed videos from train set for training and 4926 validation videos for evaluation.

\subsection{Implementation Details}
We use I3D \cite{DBLP:conf/cvpr/CarreiraZ17} network pretrained on Kinetics \cite{DBLP:conf/cvpr/CarreiraZ17} for feature extraction. The number of sampled snippets $T$ for THUMOS'14 and ActivityNet v1.3 is set to 750 and 75 respectively. The temperature parameter $\tau$ in InfoNCE loss in Equation \ref{In-Video} is set to $0.07$. $\lambda_{1}$ and $\lambda_{2}$ in Equation \ref{total} are set to 0.01 and 0.002. We train the proposal generation network for 6000 iterations with a learning rate of $1e-4$. The sizes of dilation/erosion masks $\mathcal{M}$ and $m$ are set to 6 and 3. The $k$ and $k^{hard}$ value for selecting easy and hard snippets are set to $T/5$ and $T/20$. The IoU score threshold $\gamma$ and confidence threshold $\eta$ are set to 0.2 and 0.4 respectively. As for the regression-based network, we choose TriDet architecture \cite{DBLP:conf/cvpr/ShiZCMLT23}, and train it for 6000 iterations with a learning rate of $1e-4$. We follow TriDet's original settings on all the other hyper parameters. The confidence threshold $\eta'$ for selecting proposals of EMA teacher is set to 0.4. The training batch size is 16 for THUMOS'14 and 64 for ActivityNet v1.3. All the experiments are conducted on a single Tesla V100 GPU.

\subsection{Comparison with State-of-the-Art}
\input{tables/maintable.tex}
We compare FuSTAL with several fully-supervised and weakly-supervised temporal action localization methods. Following previous works \cite{DBLP:conf/cvpr/RizveM0HSSC23}, we report the mAP over IoU thresholds 0.1:0.1:0.7 on THUMOS'14 dataset, the results are shown in Table \ref*{main}. Our FuSTAL outperforms all the previous cutting-edge methods by establishing a new state-of-the-art performance with $50.8\%$ average mAP over IoUs. Our method is also the first algorithm to reach the milestone of $50\%$ among all the WSTAL methods. FuSTAL surpasses the previous best PivoTAL, which is also a pseudo label learning method, by a large improvement of $1.2\%$. Considering the performance on the average mAP over IoU thresholds 0.1:0.1:0.5 and 0.3:0.1:0.7, FuSTAL also accomplishes excellent improvements of $1.8\%$ and $0.6\%$ respectively, reflecting the overall effectiveness. Comparing with the fully-supervised methods, the performance of our method also gets close to or even outperforms some of them.

We also conduct comparison on more difficult ActivityNet v1.3, and report the average mAP over IoU thresholds 0.50:0.05:0.95. As shown in Table \ref{t:anet}, FuSTAL obtains a result of $28.4\%$ on average mAP over IoUs, surpasses all other methods in the table.  

\subsection{Ablation Studies}
\input{tables/anet.tex}
\input{tables/ablation.tex}
To evaluate the effectiveness of each component and method in our algorithm, we conduct comprehensive ablation studies, and the results are reported in Table \ref{t:ablation}. As shown in the table, the base contrastive-based proposal generation method only achieves an average mAP of $40.9\%$. However, equipped with the cross-video contrastive loss we design, the quality of proposals witnesses a huge promotion to $43.7\%$ with an improvement of $2.8\%$. If all the proposals are provided to the regression-based model to learn at this time, it can only slightly increase to $45.0\%$, but is already $1.2\%$ higher than the performance achieved by learning the proposals without cross-video information. With simple thresholding to filter out most proposals with low confidence score, the training of regression-based model could easily accomplish an mAP of $49\%$, and continue to climb to $49.5\%$ if an MIL loss is applied to help learn discriminative features. For better proposals set with less harmful false positives, we compute inner IoU matrix and calculate the IoU score for each, filtering out a lot of useless proposals, and this procedure brings a promotion of $0.6\%$ to $50.1\%$. Finally, switching the EMA teacher to become the proposal generator makes the final improvement to $50.8\%$, which is the state-of-the-art result. A series of ablation studies on each component verify the effectiveness of them.

\input{tables/class_agnostic.tex}
During \textit{Generation-Stage} for pseudo label generation, we perform a class-agnostic contrastive learning when multiple action categories occur in one video. DCC \cite{DBLP:conf/cvpr/LiYJW022} separates classes in this case and perform contrastive learning within same predicted class. However, we conduct an ablation study to testify the rationality of our class-agnostic design in \textit{Generation-Stage} on THUMOS'14, and the results are shown in Table \ref{t:class_agnostic}. `Multi-Label Only' stands for only using examples with multi-labels, `Block Out' stands for simply block out all examples with multi-labels. `Class-Agnostic' is our strategy while `Class-Aware' refers to conducting contrast within the same category, and the category information is obtained from the prediction on each snippet. According to the results, our class-agnostic strategy achieves best results in pseudo label quality. We suggest there are several reasons for this: First, the class-aware prediction is not stable or reliable during early training, limiting the contrast within category blocks out many potential foreground snippets; Second, there are several chances for the model to correct this interference in a multi-stage framework; Third, since the video is untrimmed, different actions happen under the same scenario can help each other be distinguished from background better. 

\subsection{Qualitative Results}
\begin{figure}[t]
    \centering
    \includegraphics[width=0.75\columnwidth]{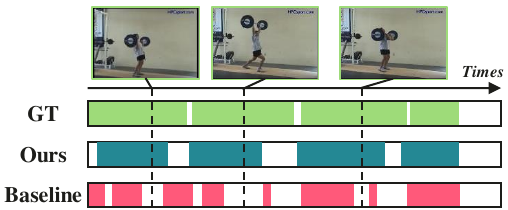}
    \vspace{-10pt}
    \caption{Qualitative comparison with ground-truth and baseline method on a `\textit{Clean and Jerk}' video. FuSTAL produces more continuous and accurate action snippets.}
    \label{fig:qualitative}
    \vspace{-10pt}
\end{figure}

We visualize the predicted action snippets in a THUMOS'14 video along the temporal axis in Figure \ref{fig:qualitative}, and compare the result with a baseline algorithm \cite{DBLP:conf/mm/HongFXSZ21}. Our algorithm is able to generate more continuous action snippets then the competitor. We believe this is due to the stronger ability to excavate core characteristics of the whole `\textit{Clean and Jerk}' action, which is consisted of a series of moves. Besides, the direct training of regression-based model further helps to view the action from a macro perspective instead of combining little snippets by classification. On the contrary, the result of baseline method is relatively scattered, and only the most discriminative parts are noticed.

\begin{figure}[t]
  \centering
  \includegraphics[width=0.6\columnwidth]{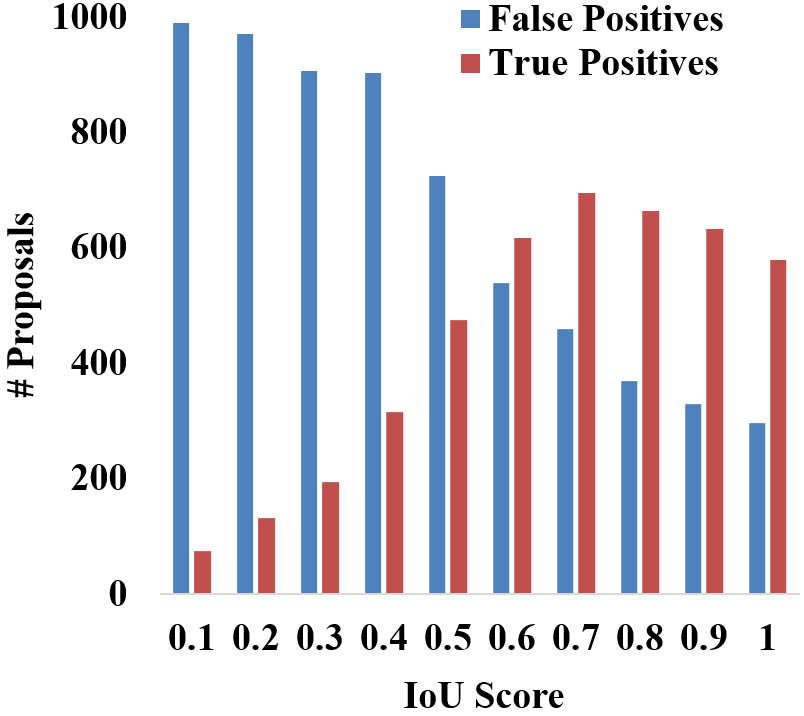}
  \vspace{-10pt}
  \caption{Distribution of false-positive/true-positive proposals number in different IoU score intervals. The blue column denotes the number of false positives with IoU score locates in $(x-0.1,x]$, and so is the red for true positives.
  }
  \label{fig:distribution}
  \vspace{-10pt} 
\end{figure}

\subsection{Discussion about Prior knowledge}
In \textit{Selection-Stage}, we quote a prior that classification-based generated proposals overlapping less with others are more likely to be the false positives which do not provide useful information about ground-truths. To further elaborate this, we visualize the distribution of generated proposals here. We here define that proposals whose maximum IoU with the ground-truths is lower than 0.1 are false positives, which can seldom provide useful information. The rest are true positive proposals. We count the number of proposals from THUMOS'14 training set belonging to these two categories with IoU scores locate in several score intervals $(x-0.1,x]$, the distribution is shown in Figure \ref{fig:distribution}. As it could be seen, the distribution shows a clear distinction between false positives and the proposals overlapping with ground-truths: most false positives have very low IoU scores, while true positives account for the majority of high score proposals. Using this observation, we are able to filter out as many harmful proposals as we could. However, the larger the threshold is not the better for model training, since more and more useful true positives would be dropped if the threshold is too high.

\section{Conclusion}
We propose a pseudo label learning framework for WSTAL problem. We explore the potential of promoting pseudo label quality at full-stage. A cross-video contrastive learning method is introduced to excavate more essential characteristics of actions at pseudo label \textit{Generation-Stage}. A \textit{Selection-Stage} is added to apply prior-knowledge to filter out useless false positives. At late \textit{Training-Stage}, student model enjoys EMA-distillation for final promotion. A series of experiments across datasets have demonstrated the effectiveness of our FuSTAL. 

\bibliographystyle{ACM-Reference-Format}
\bibliography{sample-base}

\end{document}

%% file: tables/maintable.tex
\begin{table*}[t]
    \caption{Temporal action localization perfomance compared with other state-of-the-art methods on THUMOS'14 dataset. Our algorithm outperforms all other existing algoirthms on every different IoU threshold and surpasses previous best method by 1.2\% on average mAP.}
    \centering
    \resizebox{0.98\textwidth}{!}{
    \begin{tabular}{cccccccccc|ccc}
        \toprule[2pt]
        \multirow{2}{*}{\textbf{Supervision}} & \multirow{2}{*}{\textbf{Method}} & \multirow{2}{*}{\textbf{Publication}} & \multicolumn{7}{c|}{\textbf{mAP@IoU(\%)}}       & \multicolumn{3}{c}{\textbf{AVG(\%)}}         \\
                                              &                                  &                                       & 0.1  & 0.2  & 0.3  & 0.4  & 0.5  & 0.6  & 0.7   & 0.1:0.5 & 0.3:0.7 & 0.1:0.7      \\
        \midrule[2pt]
        \multirow{4}{*}{\textbf{Full}}        & \textbf{SSN}\cite{DBLP:conf/iccv/ZhaoXWWTL17}        & ICCV 2017                     & 60.3 & 56.2 & 50.6 & 40.8 & 29.1 & -    & -    & 49.6 & -    & -            \\
                                              & \textbf{BSN}\cite{DBLP:conf/eccv/LinZSWY18}          & ECCV 2018                     & -    & -    & 53.5 & 45.0 & 36.9 & 28.4 & 20.0 & -    & 36.8 & -             \\
                                              & \textbf{TAL-Net}\cite{DBLP:conf/cvpr/ChaoVSRDS18}    & CVPR 2018                     & 59.8 & 57.1 & 53.2 & 48.5 & 42.8 & 33.8 & 20.8 & 52.3 & 39.8 & 45.1          \\
                                              & \textbf{GTAN}\cite{DBLP:conf/cvpr/LongYQTLM19}       & CVPR 2019                     & 69.1 & 63.7 & 57.8 & 47.2 & 38.8 & -    & -    & 55.3 & -    & -             \\
                                              & \textbf{TCANet}\cite{DBLP:conf/cvpr/QingSGW0W0YGS21} & CVPR 2021                     & -    & -    & 60.6 & 53.2 & 44.6 & 36.8 & 26.7 & -    & 44.4 & -             \\
        \midrule
        \multirow{18}{*}{\textbf{Weakly}}     & \textbf{HAM-Net}\cite{DBLP:conf/aaai/IslamLR21}      & AAAI 2021            & 65.4 & 59.0 & 50.3 & 41.1 & 31.0 & 20.7 & 11.1 & 49.4 & 30.8 & 39.8          \\
                                              & \textbf{CoLA}\cite{zhang2021cola}                    & CVPR 2021                     & 66.2 & 59.5 & 51.5 & 41.9 & 32.2 & 22.0 & 13.1 & 50.3 & 32.1 & 40.9          \\
                                              & \textbf{AUMN}\cite{DBLP:conf/cvpr/LuoZYL00021}       & CVPR 2021                     & 66.2 & 61.9 & 54.9 & 44.4 & 33.3 & 20.5 & 9.0  & 52.1 & 32.4 & 41.5          \\
                                              & \textbf{TS-PCA}\cite{DBLP:conf/cvpr/LiuCCDHZ21}      & CVPR 2021                     & 67.6 & 61.1 & 53.4 & 43.4 & 34.3 & 24.7 & 13.7 & 52.0 & 33.9 & 42.6          \\
                                              & \textbf{UGCT}\cite{DBLP:conf/cvpr/YangZY00021}       & CVPR 2021                     & 69.2 & 62.9 & 55.5 & 46.5 & 35.9 & 23.8 & 11.4 & 54.0 & 34.6 & 43.6          \\
                                              & \textbf{CO2-Net}\cite{DBLP:conf/mm/HongFXSZ21}       &  MM  2021                     & 70.1 & 63.6 & 54.5 & 45.7 & 38.3 & 26.4 & 13.4 & 54.4 & 35.6 & 44.6          \\
                                              & \textbf{FAC-Net}\cite{DBLP:conf/iccv/HuangWL21}      & ICCV 2021                     & 67.6 & 62.1 & 52.6 & 44.3 & 33.4 & 22.5 & 12.7 & 52.0 & 33.1 & 42.2          \\
                                              & \textbf{ACG-Net}\cite{DBLP:conf/aaai/YangQ022}       & AAAI 2022                     & 68.1 & 62.6 & 53.1 & 44.6 & 34.7 & 22.6 & 12.0 & 52.6 & 33.4 & 42.5          \\
                                              & \textbf{DCC}\cite{DBLP:conf/cvpr/LiYJW022}         & CVPR 2022                     & 69.0 & 63.8 & 55.9 & 45.9 & 35.7 & 24.3 & 13.7 & 54.1 & 35.1 & 44.0              \\
                                              & \textbf{ASM-Loc}\cite{DBLP:conf/cvpr/HeYKCZS22}      & CVPR 2022                     & 71.2 & 65.5 & 57.1 & 46.8 & 36.6 & 25.2 & 13.4 & 55.4 & 35.8 & 45.1          \\
                                              & \textbf{RSKP}\cite{DBLP:conf/cvpr/Huang0022}         & CVPR 2022                     & 71.3 & 65.3 & 55.8 & 47.5 & 38.2 & 25.4 & 12.5 & 55.6 & 35.9 & 45.1          \\
                                              & \textbf{DELU}\cite{mengyuan2022ECCV_DELU}            & ECCV 2022                     & 71.5 & 66.2 & 56.5 & 47.7 & 40.5 & 27.2 & 15.3 & 56.5 & 35.9 & 46.4          \\
                                              & \textbf{DDG-Net}\cite{DDG-Net}                       & ICCV 2023                     & 72.5 & 67.7 & 58.2 & 49.0 & 41.4 & 27.6 & 14.8 & 57.7 & 38.2 & 47.2          \\
                                              & \textbf{P-MIL}\cite{Ren_2023_CVPR}                   & CVPR 2023                     & 71.8 & 67.5 & 58.9 & 49.0 & 40.0 & 27.1 & 15.1 & 57.4 & 38.0 & 47.0          \\
                                              & \textbf{DCF}\cite{DBLP:conf/cvpr/JuZLZZC0W23}        & CVPR 2023                     & 73.5 & 68.8 & 61.5 & 53.8 & 42.0 & 29.4 & 16.8 & 59.9 & 40.7 & 49.4          \\
                                              & \textbf{PivoTAL}\cite{DBLP:conf/cvpr/RizveM0HSSC23}  & CVPR 2023                     & 74.1 & 69.6 & 61.7 & 52.1 & 42.8 & 30.6 & 16.7 & \underline{60.1} & \underline{40.8} & \underline{49.6}          \\
                                              & \textbf{ISSF}\cite{DBLP:conf/aaai/YunQWM24}  & AAAI 2024                     &72.4 &66.9 &58.4 &49.7 &41.8 &25.5 &12.8 &57.8 &37.6 &46.8          \\
                                              & \textbf{FuSTAL(Ours)}           & \textbf{-}                    
                                              & 76.6 & 71.8 & 64.0 & 54.7 & 42.2 
                                              & 30.0 & 16.2 & $\textbf{61.9}_{\uparrow1.8}$ & $\textbf{41.4}_{\uparrow0.6}$ 
                                              & $\textbf{50.8}_{\uparrow1.2}$                              \\
        \bottomrule[2pt]
        \end{tabular}
    }
    \label{main}
    \vspace{-10pt}
\end{table*}

%% file: tables/anet.tex
\begin{table}[t]
    \caption{Perfomance compared with other state-of-the-art methods on ActivityNet v1.3 dataset.}
    \centering\
    \resizebox{0.45\textwidth}{!}{
    \begin{tabular}{cccccccc}
        \toprule[2pt]
        \multirow{2}{*}{\textbf{Sup.}}        & \multirow{2}{*}{\textbf{Method}}                                          & \multicolumn{3}{c}{\textbf{mAP@IoU(\%)}}       & \multirow{2}{*}{\textbf{AVG(\%)}}         \\
                                              &                                                                           & 0.50  & 0.75 & 0.95                              &                                           \\
        \midrule[2pt]
        \multirow{2}{*}{\textbf{Full}}        & \textbf{TAL}\cite{DBLP:conf/cvpr/ChaoVSRDS18}                             & 38.2 & 18.3 & 1.3 & 20.2           \\
                                              & \textbf{BSN}\cite{DBLP:conf/eccv/LinZSWY18}                               & 46.5 & 30.0 & 8.0 & 30.0           \\
        \midrule
        \multirow{11}{*}{\textbf{Weakly}}     & \textbf{BaS-Net}\cite{DBLP:conf/aaai/LeeUB20}                              & 34.5 & 22.5 & 4.9 & 22.2           \\
                                              & \textbf{AUMN}\cite{DBLP:conf/cvpr/LuoZYL00021}                            & 38.3 & 23.5 & 5.2 & 23.5           \\
                                              & \textbf{WUM}\cite{DBLP:conf/aaai/LeeWLB21}                                & 37.0 & 23.9 & 5.7 & 23.7           \\
                                              & \textbf{UGCT}\cite{DBLP:conf/cvpr/YangZY00021}                            & 39.1 & 22.4 & 5.8 & 23.8           \\
                                              & \textbf{RSKP}\cite{DBLP:conf/cvpr/Huang0022}                              & 40.6 & 24.6 & 5.9 & 25.0           \\
                                              & \textbf{ASM-Loc}\cite{DBLP:conf/cvpr/HeYKCZS22}                           & 41.0 & 24.9 & 6.2 & 25.1           \\
                                              & \textbf{P-MIL}\cite{Ren_2023_CVPR}                                        & 41.8 & 25.4 & 5.2 & 25.5           \\
                                              & \textbf{Li et al.}\cite{Li_2023_CVPR}                                     & 41.8 & 26.0 & 6.0 & 26.0           \\
                                              & \textbf{CASE}\cite{liu2023revisiting}                                     & 43.2 & 26.2 & 6.7 & 26.8           \\
                                              & \textbf{PivoTAL}\cite{DBLP:conf/cvpr/RizveM0HSSC23}                       & 45.1 & 28.2 & 5.0 & 28.1           \\
                                              & \textbf{FuSTAL(Ours)}                                                    &45.8  &28.4  &5.1  &\textbf{28.4}            \\
        \bottomrule[2pt]
        \end{tabular}
    }
    \label{t:anet}
    \vspace{-15pt} 
\end{table}

%% file: tables/ablation.tex
\begin{table}[t]
    \caption{Ablation study results on each component.}
    \centering\
    \resizebox{0.45\textwidth}{!}{
    \begin{tabular}{ccccccc|c}
        \toprule[2pt]
        \textbf{\begin{tabular}[c]{@{}c@{}}Base\\ Method\end{tabular}} & \textbf{\begin{tabular}[c]{@{}c@{}}Cross-Video\\ Contrastive\end{tabular}} & \textbf{\begin{tabular}[c]{@{}c@{}}Regression-\\ Based Model\end{tabular}} & \textbf{\begin{tabular}[c]{@{}c@{}}Thresh\\ Out\end{tabular}} & \textbf{\begin{tabular}[c]{@{}c@{}}MIL\\ Loss\end{tabular}} & \textbf{\begin{tabular}[c]{@{}c@{}}FP\\ Filter\end{tabular}} & \textbf{\begin{tabular}[c]{@{}c@{}}EMA\\ Teacher\end{tabular}} & \textbf{mAP} \\
        \midrule[2pt]
        \checkmark &            &            &            &            &            &            & 40.9 \\
        \checkmark & \checkmark &            &            &            &            &            & 43.7 \\
        \checkmark &            & \checkmark &            &            &            &            & 43.8 \\
        \checkmark & \checkmark & \checkmark &            &            &            &            & 45.0 \\
        \checkmark & \checkmark & \checkmark & \checkmark &            &            &            & 49.0 \\
        \checkmark & \checkmark & \checkmark & \checkmark & \checkmark &            &            & 49.5 \\
        \checkmark & \checkmark & \checkmark & \checkmark & \checkmark & \checkmark &            & 50.1 \\
        \checkmark & \checkmark & \checkmark & \checkmark & \checkmark & \checkmark & \checkmark & \textbf{50.8} \\
        \bottomrule[2pt]
        \end{tabular}
    }
    \label{t:ablation}
    \vspace{-10pt}
\end{table}

%% file: tables/class_agnostic.tex
\begin{table}[t]
    \caption{Ablation experiment on class-agnostic design in contrastive learning.}
    \centering\
    \resizebox{0.45\textwidth}{!}{
    \begin{tabular}{c|ccccccc}
        \toprule[1pt]
        \textbf{\begin{tabular}[c]{@{}c@{}}Method\end{tabular}} & \textbf{\begin{tabular}[c]{@{}c@{}}Without\\ Cross-Video\end{tabular}} & \textbf{\begin{tabular}[c]{@{}c@{}}Multi-Label\\ Only\end{tabular}} & \textbf{\begin{tabular}[c]{@{}c@{}}Block\\ Out\end{tabular}} & \textbf{\begin{tabular}[c]{@{}c@{}}Class-\\Aware\end{tabular}} & \textbf{\begin{tabular}[c]{@{}c@{}}Class-\\Agnostic\end{tabular}} \\
        \midrule[1pt]
        mAP(\%) & 40.9 & 41.9 & 42.3 & 42.4 & \textbf{43.7} \\
        \bottomrule[1pt]
        \end{tabular}
    }
    \label{t:class_agnostic}  
    \vspace{-10pt} 
\end{table}

%% file: arxiv-version.bbl

\begin{thebibliography}{61}

%%% the Web address.

\ifx \showCODEN    \undefined \def \showCODEN     #1{\unskip}     \fi
\ifx \showDOI      \undefined \def \showDOI       #1{#1}\fi
\ifx \showISBNx    \undefined \def \showISBNx     #1{\unskip}     \fi
\ifx \showISBNxiii \undefined \def \showISBNxiii  #1{\unskip}     \fi
\ifx \showISSN     \undefined \def \showISSN      #1{\unskip}     \fi
\ifx \showLCCN     \undefined \def \showLCCN      #1{\unskip}     \fi
\ifx \shownote     \undefined \def \shownote      #1{#1}          \fi
\ifx \showarticletitle \undefined \def \showarticletitle #1{#1}   \fi
\ifx \showURL      \undefined \def \showURL       {\relax}        \fi
% invisible to TeX
\providecommand\bibfield[2]{#2}
\providecommand\bibinfo[2]{#2}
\providecommand\natexlab[1]{#1}
\providecommand\showeprint[2][]{arXiv:#2}

\bibitem[Carreira and Zisserman(2017)]%
        {DBLP:conf/cvpr/CarreiraZ17}
\bibfield{author}{\bibinfo{person}{Jo{\~{a}}o Carreira} {and} \bibinfo{person}{Andrew Zisserman}.} \bibinfo{year}{2017}\natexlab{}.
\newblock \showarticletitle{Quo Vadis, Action Recognition? {A} New Model and the Kinetics Dataset}. In \bibinfo{booktitle}{\emph{{IEEE} Conference on Computer Vision and Pattern Recognition, {CVPR} 2017}}. \bibinfo{publisher}{{IEEE} Computer Society}, \bibinfo{pages}{4724--4733}.
\newblock


\bibitem[Chao et~al\mbox{.}(2018)]%
        {DBLP:conf/cvpr/ChaoVSRDS18}
\bibfield{author}{\bibinfo{person}{Yu{-}Wei Chao}, \bibinfo{person}{Sudheendra Vijayanarasimhan}, \bibinfo{person}{Bryan Seybold}, \bibinfo{person}{David~A. Ross}, \bibinfo{person}{Jia Deng}, {and} \bibinfo{person}{Rahul Sukthankar}.} \bibinfo{year}{2018}\natexlab{}.
\newblock \showarticletitle{Rethinking the Faster {R-CNN} Architecture for Temporal Action Localization}. In \bibinfo{booktitle}{\emph{{IEEE} Conference on Computer Vision and Pattern Recognition, {CVPR} 2018}}. \bibinfo{pages}{1130--1139}.
\newblock


\bibitem[Chen et~al\mbox{.}(2022)]%
        {mengyuan2022ECCV_DELU}
\bibfield{author}{\bibinfo{person}{Mengyuan Chen}, \bibinfo{person}{Junyu Gao}, \bibinfo{person}{Shicai Yang}, {and} \bibinfo{person}{Changsheng Xu}.} \bibinfo{year}{2022}\natexlab{}.
\newblock \showarticletitle{Dual-Evidential Learning for Weakly-supervised Temporal Action Localization}. In \bibinfo{booktitle}{\emph{{ECCV} 2022 - 17th European Conference, Proceedings, Part {IV}}} \emph{(\bibinfo{series}{Lecture Notes in Computer Science}, Vol.~\bibinfo{volume}{13664})}. \bibinfo{pages}{192--208}.
\newblock


\bibitem[Gaidon et~al\mbox{.}(2013)]%
        {DBLP:journals/pami/GaidonHS13}
\bibfield{author}{\bibinfo{person}{Adrien Gaidon}, \bibinfo{person}{Za{\"{\i}}d Harchaoui}, {and} \bibinfo{person}{Cordelia Schmid}.} \bibinfo{year}{2013}\natexlab{}.
\newblock \showarticletitle{Temporal Localization of Actions with Actoms}.
\newblock \bibinfo{journal}{\emph{{IEEE} Trans. Pattern Anal. Mach. Intell.}} \bibinfo{volume}{35}, \bibinfo{number}{11} (\bibinfo{year}{2013}), \bibinfo{pages}{2782--2795}.
\newblock


\bibitem[He et~al\mbox{.}(2022)]%
        {DBLP:conf/cvpr/HeYKCZS22}
\bibfield{author}{\bibinfo{person}{Bo He}, \bibinfo{person}{Xitong Yang}, \bibinfo{person}{Le Kang}, \bibinfo{person}{Zhiyu Cheng}, \bibinfo{person}{Xin Zhou}, {and} \bibinfo{person}{Abhinav Shrivastava}.} \bibinfo{year}{2022}\natexlab{}.
\newblock \showarticletitle{ASM-Loc: Action-aware Segment Modeling for Weakly-Supervised Temporal Action Localization}. In \bibinfo{booktitle}{\emph{{IEEE/CVF} Conference on Computer Vision and Pattern Recognition, {CVPR} 2022}}. \bibinfo{pages}{13915--13925}.
\newblock


\bibitem[He et~al\mbox{.}(2020)]%
        {DBLP:conf/cvpr/He0WXG20}
\bibfield{author}{\bibinfo{person}{Kaiming He}, \bibinfo{person}{Haoqi Fan}, \bibinfo{person}{Yuxin Wu}, \bibinfo{person}{Saining Xie}, {and} \bibinfo{person}{Ross~B. Girshick}.} \bibinfo{year}{2020}\natexlab{}.
\newblock \showarticletitle{Momentum Contrast for Unsupervised Visual Representation Learning}. In \bibinfo{booktitle}{\emph{{IEEE/CVF} Conference on Computer Vision and Pattern Recognition, {CVPR} 2020}}. \bibinfo{pages}{9726--9735}.
\newblock


\bibitem[Heilbron et~al\mbox{.}(2015)]%
        {DBLP:conf/cvpr/HeilbronEGN15}
\bibfield{author}{\bibinfo{person}{Fabian~Caba Heilbron}, \bibinfo{person}{Victor Escorcia}, \bibinfo{person}{Bernard Ghanem}, {and} \bibinfo{person}{Juan~Carlos Niebles}.} \bibinfo{year}{2015}\natexlab{}.
\newblock \showarticletitle{ActivityNet: {A} large-scale video benchmark for human activity understanding}. In \bibinfo{booktitle}{\emph{{IEEE} Conference on Computer Vision and Pattern Recognition, {CVPR} 2015}}. \bibinfo{publisher}{{IEEE} Computer Society}, \bibinfo{pages}{961--970}.
\newblock


\bibitem[Hong et~al\mbox{.}(2021)]%
        {DBLP:conf/mm/HongFXSZ21}
\bibfield{author}{\bibinfo{person}{Fa{-}Ting Hong}, \bibinfo{person}{Jia{-}Chang Feng}, \bibinfo{person}{Dan Xu}, \bibinfo{person}{Ying Shan}, {and} \bibinfo{person}{Wei{-}Shi Zheng}.} \bibinfo{year}{2021}\natexlab{}.
\newblock \showarticletitle{Cross-modal Consensus Network for Weakly Supervised Temporal Action Localization}. In \bibinfo{booktitle}{\emph{{MM} '21: {ACM} Multimedia Conference}}. \bibinfo{pages}{1591--1599}.
\newblock


\bibitem[Huang et~al\mbox{.}(2021)]%
        {DBLP:conf/iccv/HuangWL21}
\bibfield{author}{\bibinfo{person}{Linjiang Huang}, \bibinfo{person}{Liang Wang}, {and} \bibinfo{person}{Hongsheng Li}.} \bibinfo{year}{2021}\natexlab{}.
\newblock \showarticletitle{Foreground-Action Consistency Network for Weakly Supervised Temporal Action Localization}. In \bibinfo{booktitle}{\emph{2021 {IEEE/CVF} International Conference on Computer Vision, {ICCV} 2021}}. \bibinfo{pages}{7982--7991}.
\newblock


\bibitem[Huang et~al\mbox{.}(2022)]%
        {DBLP:conf/cvpr/Huang0022}
\bibfield{author}{\bibinfo{person}{Linjiang Huang}, \bibinfo{person}{Liang Wang}, {and} \bibinfo{person}{Hongsheng Li}.} \bibinfo{year}{2022}\natexlab{}.
\newblock \showarticletitle{Weakly Supervised Temporal Action Localization via Representative Snippet Knowledge Propagation}. In \bibinfo{booktitle}{\emph{{IEEE/CVF} Conference on Computer Vision and Pattern Recognition, {CVPR} 2022}}. \bibinfo{pages}{3262--3271}.
\newblock


\bibitem[Idrees et~al\mbox{.}(2017)]%
        {DBLP:journals/cviu/IdreesZJGLSS17}
\bibfield{author}{\bibinfo{person}{Haroon Idrees}, \bibinfo{person}{Amir~R. Zamir}, \bibinfo{person}{Yu{-}Gang Jiang}, \bibinfo{person}{Alex Gorban}, \bibinfo{person}{Ivan Laptev}, \bibinfo{person}{Rahul Sukthankar}, {and} \bibinfo{person}{Mubarak Shah}.} \bibinfo{year}{2017}\natexlab{}.
\newblock \showarticletitle{The {THUMOS} challenge on action recognition for videos "in the wild"}.
\newblock \bibinfo{journal}{\emph{Comput. Vis. Image Underst.}}  \bibinfo{volume}{155} (\bibinfo{year}{2017}), \bibinfo{pages}{1--23}.
\newblock


\bibitem[Islam et~al\mbox{.}(2021)]%
        {DBLP:conf/aaai/IslamLR21}
\bibfield{author}{\bibinfo{person}{Ashraful Islam}, \bibinfo{person}{Chengjiang Long}, {and} \bibinfo{person}{Richard~J. Radke}.} \bibinfo{year}{2021}\natexlab{}.
\newblock \showarticletitle{A Hybrid Attention Mechanism for Weakly-Supervised Temporal Action Localization}. In \bibinfo{booktitle}{\emph{Thirty-Fifth {AAAI} Conference on Artificial Intelligence, {AAAI} 2021}}. \bibinfo{pages}{1637--1645}.
\newblock


\bibitem[Ju et~al\mbox{.}(2023)]%
        {DBLP:conf/cvpr/JuZLZZC0W23}
\bibfield{author}{\bibinfo{person}{Chen Ju}, \bibinfo{person}{Kunhao Zheng}, \bibinfo{person}{Jinxiang Liu}, \bibinfo{person}{Peisen Zhao}, \bibinfo{person}{Ya Zhang}, \bibinfo{person}{Jianlong Chang}, \bibinfo{person}{Qi Tian}, {and} \bibinfo{person}{Yanfeng Wang}.} \bibinfo{year}{2023}\natexlab{}.
\newblock \showarticletitle{Distilling Vision-Language Pre-Training to Collaborate with Weakly-Supervised Temporal Action Localization}. In \bibinfo{booktitle}{\emph{{IEEE/CVF} Conference on Computer Vision and Pattern Recognition, {CVPR} 2023}}. \bibinfo{pages}{14751--14762}.
\newblock


\bibitem[Lee et~al\mbox{.}(2013)]%
        {lee2013pseudo}
\bibfield{author}{\bibinfo{person}{Dong-Hyun Lee} {et~al\mbox{.}}} \bibinfo{year}{2013}\natexlab{}.
\newblock \showarticletitle{Pseudo-label: The simple and efficient semi-supervised learning method for deep neural networks}. In \bibinfo{booktitle}{\emph{Workshop on challenges in representation learning, ICML}}, Vol.~\bibinfo{volume}{3}. Atlanta, \bibinfo{pages}{896}.
\newblock


\bibitem[Lee et~al\mbox{.}(2020)]%
        {DBLP:conf/aaai/LeeUB20}
\bibfield{author}{\bibinfo{person}{Pilhyeon Lee}, \bibinfo{person}{Youngjung Uh}, {and} \bibinfo{person}{Hyeran Byun}.} \bibinfo{year}{2020}\natexlab{}.
\newblock \showarticletitle{Background Suppression Network for Weakly-Supervised Temporal Action Localization}. In \bibinfo{booktitle}{\emph{The Thirty-Fourth {AAAI} Conference on Artificial Intelligence, {AAAI} 2020}}. \bibinfo{pages}{11320--11327}.
\newblock


\bibitem[Lee et~al\mbox{.}(2021)]%
        {DBLP:conf/aaai/LeeWLB21}
\bibfield{author}{\bibinfo{person}{Pilhyeon Lee}, \bibinfo{person}{Jinglu Wang}, \bibinfo{person}{Yan Lu}, {and} \bibinfo{person}{Hyeran Byun}.} \bibinfo{year}{2021}\natexlab{}.
\newblock \showarticletitle{Weakly-supervised Temporal Action Localization by Uncertainty Modeling}. In \bibinfo{booktitle}{\emph{Thirty-Fifth {AAAI} Conference on Artificial Intelligence, {AAAI} 2021}}. \bibinfo{publisher}{{AAAI} Press}, \bibinfo{pages}{1854--1862}.
\newblock


\bibitem[Lee et~al\mbox{.}(2012)]%
        {DBLP:conf/cvpr/LeeGG12}
\bibfield{author}{\bibinfo{person}{Yong~Jae Lee}, \bibinfo{person}{Joydeep Ghosh}, {and} \bibinfo{person}{Kristen Grauman}.} \bibinfo{year}{2012}\natexlab{}.
\newblock \showarticletitle{Discovering important people and objects for egocentric video summarization}. In \bibinfo{booktitle}{\emph{{IEEE} Conference on Computer Vision and Pattern Recognition, {CVPR} 2012}}. \bibinfo{pages}{1346--1353}.
\newblock


\bibitem[Li et~al\mbox{.}(2023)]%
        {Li_2023_CVPR}
\bibfield{author}{\bibinfo{person}{Guozhang Li}, \bibinfo{person}{De Cheng}, \bibinfo{person}{Xinpeng Ding}, \bibinfo{person}{Nannan Wang}, \bibinfo{person}{Xiaoyu Wang}, {and} \bibinfo{person}{Xinbo Gao}.} \bibinfo{year}{2023}\natexlab{}.
\newblock \showarticletitle{Boosting Weakly-Supervised Temporal Action Localization With Text Information}. In \bibinfo{booktitle}{\emph{Proceedings of the IEEE/CVF Conference on Computer Vision and Pattern Recognition (CVPR)}}. \bibinfo{pages}{10648--10657}.
\newblock


\bibitem[Li et~al\mbox{.}(2022)]%
        {DBLP:conf/cvpr/LiYJW022}
\bibfield{author}{\bibinfo{person}{Jingjing Li}, \bibinfo{person}{Tianyu Yang}, \bibinfo{person}{Wei Ji}, \bibinfo{person}{Jue Wang}, {and} \bibinfo{person}{Li Cheng}.} \bibinfo{year}{2022}\natexlab{}.
\newblock \showarticletitle{Exploring Denoised Cross-video Contrast for Weakly-supervised Temporal Action Localization}. In \bibinfo{booktitle}{\emph{{IEEE/CVF} Conference on Computer Vision and Pattern Recognition, {CVPR} 2022}}. \bibinfo{publisher}{{IEEE}}, \bibinfo{pages}{19882--19892}.
\newblock


\bibitem[Lin et~al\mbox{.}(2021)]%
        {DBLP:conf/cvpr/Lin0LWTWLHF21}
\bibfield{author}{\bibinfo{person}{Chuming Lin}, \bibinfo{person}{Chengming Xu}, \bibinfo{person}{Donghao Luo}, \bibinfo{person}{Yabiao Wang}, \bibinfo{person}{Ying Tai}, \bibinfo{person}{Chengjie Wang}, \bibinfo{person}{Jilin Li}, \bibinfo{person}{Feiyue Huang}, {and} \bibinfo{person}{Yanwei Fu}.} \bibinfo{year}{2021}\natexlab{}.
\newblock \showarticletitle{Learning Salient Boundary Feature for Anchor-free Temporal Action Localization}. In \bibinfo{booktitle}{\emph{{IEEE} Conference on Computer Vision and Pattern Recognition, {CVPR} 2021}}. \bibinfo{pages}{3320--3329}.
\newblock


\bibitem[Lin et~al\mbox{.}(2017)]%
        {DBLP:conf/iccv/LinGGHD17}
\bibfield{author}{\bibinfo{person}{Tsung{-}Yi Lin}, \bibinfo{person}{Priya Goyal}, \bibinfo{person}{Ross~B. Girshick}, \bibinfo{person}{Kaiming He}, {and} \bibinfo{person}{Piotr Doll{\'{a}}r}.} \bibinfo{year}{2017}\natexlab{}.
\newblock \showarticletitle{Focal Loss for Dense Object Detection}. In \bibinfo{booktitle}{\emph{{IEEE} International Conference on Computer Vision, {ICCV} 2017}}. \bibinfo{publisher}{{IEEE} Computer Society}, \bibinfo{pages}{2999--3007}.
\newblock


\bibitem[Lin et~al\mbox{.}(2018)]%
        {DBLP:conf/eccv/LinZSWY18}
\bibfield{author}{\bibinfo{person}{Tianwei Lin}, \bibinfo{person}{Xu Zhao}, \bibinfo{person}{Haisheng Su}, \bibinfo{person}{Chongjing Wang}, {and} \bibinfo{person}{Ming Yang}.} \bibinfo{year}{2018}\natexlab{}.
\newblock \showarticletitle{{BSN:} Boundary Sensitive Network for Temporal Action Proposal Generation}. In \bibinfo{booktitle}{\emph{{ECCV} 2018 - 15th European Conference, Proceedings, Part {IV}}} \emph{(\bibinfo{series}{Lecture Notes in Computer Science}, Vol.~\bibinfo{volume}{11208})}. \bibinfo{pages}{3--21}.
\newblock


\bibitem[Liu et~al\mbox{.}(2023)]%
        {liu2023revisiting}
\bibfield{author}{\bibinfo{person}{Qinying Liu}, \bibinfo{person}{Zilei Wang}, \bibinfo{person}{Shenghai Rong}, \bibinfo{person}{Junjie Li}, {and} \bibinfo{person}{Yixin Zhang}.} \bibinfo{year}{2023}\natexlab{}.
\newblock \showarticletitle{Revisiting Foreground and Background Separation in Weakly-supervised Temporal Action Localization: A Clustering-based Approach}. In \bibinfo{booktitle}{\emph{Proceedings of the IEEE/CVF International Conference on Computer Vision}}. \bibinfo{pages}{10433--10443}.
\newblock


\bibitem[Liu et~al\mbox{.}(2021)]%
        {DBLP:conf/cvpr/LiuCCDHZ21}
\bibfield{author}{\bibinfo{person}{Yuan Liu}, \bibinfo{person}{Jingyuan Chen}, \bibinfo{person}{Zhenfang Chen}, \bibinfo{person}{Bing Deng}, \bibinfo{person}{Jianqiang Huang}, {and} \bibinfo{person}{Hanwang Zhang}.} \bibinfo{year}{2021}\natexlab{}.
\newblock \showarticletitle{The Blessings of Unlabeled Background in Untrimmed Videos}. In \bibinfo{booktitle}{\emph{{IEEE} Conference on Computer Vision and Pattern Recognition, {CVPR} 2021}}. \bibinfo{pages}{6176--6185}.
\newblock


\bibitem[Long et~al\mbox{.}(2019)]%
        {DBLP:conf/cvpr/LongYQTLM19}
\bibfield{author}{\bibinfo{person}{Fuchen Long}, \bibinfo{person}{Ting Yao}, \bibinfo{person}{Zhaofan Qiu}, \bibinfo{person}{Xinmei Tian}, \bibinfo{person}{Jiebo Luo}, {and} \bibinfo{person}{Tao Mei}.} \bibinfo{year}{2019}\natexlab{}.
\newblock \showarticletitle{Gaussian Temporal Awareness Networks for Action Localization}. In \bibinfo{booktitle}{\emph{{IEEE} Conference on Computer Vision and Pattern Recognition, {CVPR} 2019}}. \bibinfo{pages}{344--353}.
\newblock


\bibitem[Luo et~al\mbox{.}(2021)]%
        {DBLP:conf/cvpr/LuoZYL00021}
\bibfield{author}{\bibinfo{person}{Wang Luo}, \bibinfo{person}{Tianzhu Zhang}, \bibinfo{person}{Wenfei Yang}, \bibinfo{person}{Jingen Liu}, \bibinfo{person}{Tao Mei}, \bibinfo{person}{Feng Wu}, {and} \bibinfo{person}{Yongdong Zhang}.} \bibinfo{year}{2021}\natexlab{}.
\newblock \showarticletitle{Action Unit Memory Network for Weakly Supervised Temporal Action Localization}. In \bibinfo{booktitle}{\emph{{IEEE} Conference on Computer Vision and Pattern Recognition, {CVPR} 2021}}. \bibinfo{pages}{9969--9979}.
\newblock


\bibitem[Luo et~al\mbox{.}(2020)]%
        {luo2020weakly}
\bibfield{author}{\bibinfo{person}{Zhekun Luo}, \bibinfo{person}{Devin Guillory}, \bibinfo{person}{Baifeng Shi}, \bibinfo{person}{Wei Ke}, \bibinfo{person}{Fang Wan}, \bibinfo{person}{Trevor Darrell}, {and} \bibinfo{person}{Huijuan Xu}.} \bibinfo{year}{2020}\natexlab{}.
\newblock \showarticletitle{Weakly-supervised action localization with expectation-maximization multi-instance learning}. In \bibinfo{booktitle}{\emph{ECCV 2020: 16th European Conference, Proceedings, Part XXIX 16}}. Springer, \bibinfo{pages}{729--745}.
\newblock


\bibitem[Ma et~al\mbox{.}(2021)]%
        {DBLP:conf/cvpr/MaGVY21}
\bibfield{author}{\bibinfo{person}{Junwei Ma}, \bibinfo{person}{Satya~Krishna Gorti}, \bibinfo{person}{Maksims Volkovs}, {and} \bibinfo{person}{Guang~Wei Yu}.} \bibinfo{year}{2021}\natexlab{}.
\newblock \showarticletitle{Weakly Supervised Action Selection Learning in Video}. In \bibinfo{booktitle}{\emph{{IEEE} Conference on Computer Vision and Pattern Recognition, {CVPR} 2021}}. \bibinfo{pages}{7587--7596}.
\newblock


\bibitem[Moniruzzaman et~al\mbox{.}(2020)]%
        {DBLP:conf/mm/MoniruzzamanYHQ20}
\bibfield{author}{\bibinfo{person}{Md. Moniruzzaman}, \bibinfo{person}{Zhaozheng Yin}, \bibinfo{person}{Zhihai He}, \bibinfo{person}{Ruwen Qin}, {and} \bibinfo{person}{Ming~C. Leu}.} \bibinfo{year}{2020}\natexlab{}.
\newblock \showarticletitle{Action Completeness Modeling with Background Aware Networks for Weakly-Supervised Temporal Action Localization}. In \bibinfo{booktitle}{\emph{{MM} '20: The 28th {ACM} International Conference on Multimedia}}. \bibinfo{pages}{2166--2174}.
\newblock


\bibitem[Paul et~al\mbox{.}(2018)]%
        {DBLP:conf/eccv/PaulRR18}
\bibfield{author}{\bibinfo{person}{Sujoy Paul}, \bibinfo{person}{Sourya Roy}, {and} \bibinfo{person}{Amit~K. Roy{-}Chowdhury}.} \bibinfo{year}{2018}\natexlab{}.
\newblock \showarticletitle{{W-TALC:} Weakly-Supervised Temporal Activity Localization and Classification}. In \bibinfo{booktitle}{\emph{{ECCV} 2018 - 15th European Conference, Proceedings, Part {IV}}}, Vol.~\bibinfo{volume}{11208}. \bibinfo{pages}{588--607}.
\newblock


\bibitem[Qing et~al\mbox{.}(2021)]%
        {DBLP:conf/cvpr/QingSGW0W0YGS21}
\bibfield{author}{\bibinfo{person}{Zhiwu Qing}, \bibinfo{person}{Haisheng Su}, \bibinfo{person}{Weihao Gan}, \bibinfo{person}{Dongliang Wang}, \bibinfo{person}{Wei Wu}, \bibinfo{person}{Xiang Wang}, \bibinfo{person}{Yu Qiao}, \bibinfo{person}{Junjie Yan}, \bibinfo{person}{Changxin Gao}, {and} \bibinfo{person}{Nong Sang}.} \bibinfo{year}{2021}\natexlab{}.
\newblock \showarticletitle{Temporal Context Aggregation Network for Temporal Action Proposal Refinement}. In \bibinfo{booktitle}{\emph{{IEEE} Conference on Computer Vision and Pattern Recognition, {CVPR} 2021}}. \bibinfo{pages}{485--494}.
\newblock


\bibitem[Qu et~al\mbox{.}(2021)]%
        {DBLP:journals/corr/abs-2104-02967}
\bibfield{author}{\bibinfo{person}{Sanqing Qu}, \bibinfo{person}{Guang Chen}, \bibinfo{person}{Zhijun Li}, \bibinfo{person}{Lijun Zhang}, \bibinfo{person}{Fan Lu}, {and} \bibinfo{person}{Alois~C. Knoll}.} \bibinfo{year}{2021}\natexlab{}.
\newblock \showarticletitle{ACM-Net: Action Context Modeling Network for Weakly-Supervised Temporal Action Localization}.
\newblock \bibinfo{journal}{\emph{CoRR}}  \bibinfo{volume}{abs/2104.02967} (\bibinfo{year}{2021}).
\newblock


\bibitem[Ren et~al\mbox{.}(2023)]%
        {Ren_2023_CVPR}
\bibfield{author}{\bibinfo{person}{Huan Ren}, \bibinfo{person}{Wenfei Yang}, \bibinfo{person}{Tianzhu Zhang}, {and} \bibinfo{person}{Yongdong Zhang}.} \bibinfo{year}{2023}\natexlab{}.
\newblock \showarticletitle{Proposal-Based Multiple Instance Learning for Weakly-Supervised Temporal Action Localization}. In \bibinfo{booktitle}{\emph{{IEEE/CVF} Conference on Computer Vision and Pattern Recognition, {CVPR} 2023}}. \bibinfo{pages}{2394--2404}.
\newblock


\bibitem[Rizve et~al\mbox{.}(2023)]%
        {DBLP:conf/cvpr/RizveM0HSSC23}
\bibfield{author}{\bibinfo{person}{Mamshad~Nayeem Rizve}, \bibinfo{person}{Gaurav Mittal}, \bibinfo{person}{Ye Yu}, \bibinfo{person}{Matthew Hall}, \bibinfo{person}{Sandra Sajeev}, \bibinfo{person}{Mubarak Shah}, {and} \bibinfo{person}{Mei Chen}.} \bibinfo{year}{2023}\natexlab{}.
\newblock \showarticletitle{PivoTAL: Prior-Driven Supervision for Weakly-Supervised Temporal Action Localization}. In \bibinfo{booktitle}{\emph{{IEEE/CVF} Conference on Computer Vision and Pattern Recognition, {CVPR} 2023}}. \bibinfo{pages}{22992--23002}.
\newblock


\bibitem[Shi et~al\mbox{.}(2020)]%
        {DBLP:conf/cvpr/ShiDMW20}
\bibfield{author}{\bibinfo{person}{Baifeng Shi}, \bibinfo{person}{Qi Dai}, \bibinfo{person}{Yadong Mu}, {and} \bibinfo{person}{Jingdong Wang}.} \bibinfo{year}{2020}\natexlab{}.
\newblock \showarticletitle{Weakly-Supervised Action Localization by Generative Attention Modeling}. In \bibinfo{booktitle}{\emph{{IEEE/CVF} Conference on Computer Vision and Pattern Recognition, {CVPR} 2020}}. \bibinfo{pages}{1006--1016}.
\newblock


\bibitem[Shi et~al\mbox{.}(2023)]%
        {DBLP:conf/cvpr/ShiZCMLT23}
\bibfield{author}{\bibinfo{person}{Dingfeng Shi}, \bibinfo{person}{Yujie Zhong}, \bibinfo{person}{Qiong Cao}, \bibinfo{person}{Lin Ma}, \bibinfo{person}{Jia Li}, {and} \bibinfo{person}{Dacheng Tao}.} \bibinfo{year}{2023}\natexlab{}.
\newblock \showarticletitle{TriDet: Temporal Action Detection with Relative Boundary Modeling}. In \bibinfo{booktitle}{\emph{{IEEE/CVF} Conference on Computer Vision and Pattern Recognition, {CVPR} 2023}}. \bibinfo{pages}{18857--18866}.
\newblock


\bibitem[Shou et~al\mbox{.}(2016)]%
        {DBLP:conf/cvpr/ShouWC16}
\bibfield{author}{\bibinfo{person}{Zheng Shou}, \bibinfo{person}{Dongang Wang}, {and} \bibinfo{person}{Shih{-}Fu Chang}.} \bibinfo{year}{2016}\natexlab{}.
\newblock \showarticletitle{Temporal Action Localization in Untrimmed Videos via Multi-stage CNNs}. In \bibinfo{booktitle}{\emph{{IEEE} Conference on Computer Vision and Pattern Recognition, {CVPR} 2016}}. \bibinfo{pages}{1049--1058}.
\newblock


\bibitem[Sohn et~al\mbox{.}(2020)]%
        {DBLP:conf/nips/SohnBCZZRCKL20}
\bibfield{author}{\bibinfo{person}{Kihyuk Sohn}, \bibinfo{person}{David Berthelot}, \bibinfo{person}{Nicholas Carlini}, \bibinfo{person}{Zizhao Zhang}, \bibinfo{person}{Han Zhang}, \bibinfo{person}{Colin Raffel}, \bibinfo{person}{Ekin~Dogus Cubuk}, \bibinfo{person}{Alexey Kurakin}, {and} \bibinfo{person}{Chun{-}Liang Li}.} \bibinfo{year}{2020}\natexlab{}.
\newblock \showarticletitle{FixMatch: Simplifying Semi-Supervised Learning with Consistency and Confidence}. In \bibinfo{booktitle}{\emph{Advances in Neural Information Processing Systems 33: Annual Conference on Neural Information Processing Systems 2020, NeurIPS 2020}}.
\newblock


\bibitem[Sridhar et~al\mbox{.}(2021)]%
        {DBLP:conf/iccv/SridharQMLDL21}
\bibfield{author}{\bibinfo{person}{Deepak Sridhar}, \bibinfo{person}{Niamul Quader}, \bibinfo{person}{Srikanth Muralidharan}, \bibinfo{person}{Yaoxin Li}, \bibinfo{person}{Peng Dai}, {and} \bibinfo{person}{Juwei Lu}.} \bibinfo{year}{2021}\natexlab{}.
\newblock \showarticletitle{Class Semantics-based Attention for Action Detection}. In \bibinfo{booktitle}{\emph{{IEEE/CVF} International Conference on Computer Vision, {ICCV} 2021}}. \bibinfo{pages}{13719--13728}.
\newblock


\bibitem[Tang et~al\mbox{.}(2023)]%
        {DDG-Net}
\bibfield{author}{\bibinfo{person}{Xiaojun Tang}, \bibinfo{person}{Junsong Fan}, \bibinfo{person}{Chuanchen Luo}, \bibinfo{person}{Zhaoxiang Zhang}, \bibinfo{person}{Man Zhang}, {and} \bibinfo{person}{Zongyuan Yang}.} \bibinfo{year}{2023}\natexlab{}.
\newblock \showarticletitle{DDG-Net: Discriminability-Driven Graph Network for Weakly-supervised Temporal Action Localization}.
\newblock \bibinfo{journal}{\emph{CoRR}}  \bibinfo{volume}{abs/2307.16415} (\bibinfo{year}{2023}).
\newblock


\bibitem[Vishwakarma and Agrawal(2013)]%
        {DBLP:journals/vc/VishwakarmaA13}
\bibfield{author}{\bibinfo{person}{Sarvesh Vishwakarma} {and} \bibinfo{person}{Anupam Agrawal}.} \bibinfo{year}{2013}\natexlab{}.
\newblock \showarticletitle{A survey on activity recognition and behavior understanding in video surveillance}.
\newblock \bibinfo{journal}{\emph{Vis. Comput.}} \bibinfo{volume}{29}, \bibinfo{number}{10} (\bibinfo{year}{2013}), \bibinfo{pages}{983--1009}.
\newblock


\bibitem[Wang et~al\mbox{.}(2017)]%
        {DBLP:conf/cvpr/WangXLG17}
\bibfield{author}{\bibinfo{person}{Limin Wang}, \bibinfo{person}{Yuanjun Xiong}, \bibinfo{person}{Dahua Lin}, {and} \bibinfo{person}{Luc~Van Gool}.} \bibinfo{year}{2017}\natexlab{}.
\newblock \showarticletitle{UntrimmedNets for Weakly Supervised Action Recognition and Detection}. In \bibinfo{booktitle}{\emph{{IEEE} Conference on Computer Vision and Pattern Recognition, {CVPR} 2017}}. \bibinfo{pages}{6402--6411}.
\newblock


\bibitem[Weng et~al\mbox{.}(2022)]%
        {DBLP:conf/eccv/WengPHCZ22}
\bibfield{author}{\bibinfo{person}{Yuetian Weng}, \bibinfo{person}{Zizheng Pan}, \bibinfo{person}{Mingfei Han}, \bibinfo{person}{Xiaojun Chang}, {and} \bibinfo{person}{Bohan Zhuang}.} \bibinfo{year}{2022}\natexlab{}.
\newblock \showarticletitle{An Efficient Spatio-Temporal Pyramid Transformer for Action Detection}. In \bibinfo{booktitle}{\emph{{ECCV} 2022 - 17th European Conference, Proceedings, Part {XXXIV}}} \emph{(\bibinfo{series}{Lecture Notes in Computer Science}, Vol.~\bibinfo{volume}{13694})}. \bibinfo{pages}{358--375}.
\newblock


\bibitem[Xu et~al\mbox{.}(2017)]%
        {DBLP:conf/iccv/XuDS17}
\bibfield{author}{\bibinfo{person}{Huijuan Xu}, \bibinfo{person}{Abir Das}, {and} \bibinfo{person}{Kate Saenko}.} \bibinfo{year}{2017}\natexlab{}.
\newblock \showarticletitle{{R-C3D:} Region Convolutional 3D Network for Temporal Activity Detection}. In \bibinfo{booktitle}{\emph{{IEEE} International Conference on Computer Vision, {ICCV} 2017}}. \bibinfo{pages}{5794--5803}.
\newblock


\bibitem[Xu et~al\mbox{.}(2020)]%
        {DBLP:conf/cvpr/XuZRTG20}
\bibfield{author}{\bibinfo{person}{Mengmeng Xu}, \bibinfo{person}{Chen Zhao}, \bibinfo{person}{David~S. Rojas}, \bibinfo{person}{Ali~K. Thabet}, {and} \bibinfo{person}{Bernard Ghanem}.} \bibinfo{year}{2020}\natexlab{}.
\newblock \showarticletitle{{G-TAD:} Sub-Graph Localization for Temporal Action Detection}. In \bibinfo{booktitle}{\emph{{IEEE/CVF} Conference on Computer Vision and Pattern Recognition, {CVPR} 2020}}. \bibinfo{pages}{10153--10162}.
\newblock


\bibitem[Xu et~al\mbox{.}(2019)]%
        {DBLP:conf/aaai/XuZCXNPW19}
\bibfield{author}{\bibinfo{person}{Yunlu Xu}, \bibinfo{person}{Chengwei Zhang}, \bibinfo{person}{Zhanzhan Cheng}, \bibinfo{person}{Jianwen Xie}, \bibinfo{person}{Yi Niu}, \bibinfo{person}{Shiliang Pu}, {and} \bibinfo{person}{Fei Wu}.} \bibinfo{year}{2019}\natexlab{}.
\newblock \showarticletitle{Segregated Temporal Assembly Recurrent Networks for Weakly Supervised Multiple Action Detection}. In \bibinfo{booktitle}{\emph{The Thirty-Third {AAAI} Conference on Artificial Intelligence, {AAAI} 2019}}. \bibinfo{pages}{9070--9078}.
\newblock


\bibitem[Yang et~al\mbox{.}(2020)]%
        {DBLP:journals/tip/YangPZFH20}
\bibfield{author}{\bibinfo{person}{Le Yang}, \bibinfo{person}{Houwen Peng}, \bibinfo{person}{Dingwen Zhang}, \bibinfo{person}{Jianlong Fu}, {and} \bibinfo{person}{Junwei Han}.} \bibinfo{year}{2020}\natexlab{}.
\newblock \showarticletitle{Revisiting Anchor Mechanisms for Temporal Action Localization}.
\newblock \bibinfo{journal}{\emph{{IEEE} Trans. Image Process.}}  \bibinfo{volume}{29} (\bibinfo{year}{2020}), \bibinfo{pages}{8535--8548}.
\newblock


\bibitem[Yang et~al\mbox{.}(2023)]%
        {DBLP:journals/cviu/YangCZLW23}
\bibfield{author}{\bibinfo{person}{Min Yang}, \bibinfo{person}{Guo Chen}, \bibinfo{person}{Yin{-}Dong Zheng}, \bibinfo{person}{Tong Lu}, {and} \bibinfo{person}{Limin Wang}.} \bibinfo{year}{2023}\natexlab{}.
\newblock \showarticletitle{BasicTAD: An astounding RGB-Only baseline for temporal action detection}.
\newblock \bibinfo{journal}{\emph{Comput. Vis. Image Underst.}}  \bibinfo{volume}{232} (\bibinfo{year}{2023}), \bibinfo{pages}{103692}.
\newblock


\bibitem[Yang et~al\mbox{.}(2021)]%
        {DBLP:conf/cvpr/YangZY00021}
\bibfield{author}{\bibinfo{person}{Wenfei Yang}, \bibinfo{person}{Tianzhu Zhang}, \bibinfo{person}{Xiaoyuan Yu}, \bibinfo{person}{Qi Tian}, \bibinfo{person}{Yongdong Zhang}, {and} \bibinfo{person}{Feng Wu}.} \bibinfo{year}{2021}\natexlab{}.
\newblock \showarticletitle{Uncertainty Guided Collaborative Training for Weakly Supervised Temporal Action Detection}. In \bibinfo{booktitle}{\emph{{IEEE} Conference on Computer Vision and Pattern Recognition, {CVPR} 2021}}. \bibinfo{pages}{53--63}.
\newblock


\bibitem[Yang et~al\mbox{.}(2022)]%
        {DBLP:conf/aaai/YangQ022}
\bibfield{author}{\bibinfo{person}{Zichen Yang}, \bibinfo{person}{Jie Qin}, {and} \bibinfo{person}{Di Huang}.} \bibinfo{year}{2022}\natexlab{}.
\newblock \showarticletitle{ACGNet: Action Complement Graph Network for Weakly-Supervised Temporal Action Localization}. In \bibinfo{booktitle}{\emph{Thirty-Sixth {AAAI} Conference on Artificial Intelligence, {AAAI} 2022}}. \bibinfo{pages}{3090--3098}.
\newblock


\bibitem[Yun et~al\mbox{.}(2024)]%
        {DBLP:conf/aaai/YunQWM24}
\bibfield{author}{\bibinfo{person}{Wulian Yun}, \bibinfo{person}{Mengshi Qi}, \bibinfo{person}{Chuanming Wang}, {and} \bibinfo{person}{Huadong Ma}.} \bibinfo{year}{2024}\natexlab{}.
\newblock \showarticletitle{Weakly-Supervised Temporal Action Localization by Inferring Salient Snippet-Feature}. In \bibinfo{booktitle}{\emph{Thirty-Eighth {AAAI} Conference on Artificial Intelligence, {AAAI} 2024}}. \bibinfo{publisher}{{AAAI} Press}, \bibinfo{pages}{6908--6916}.
\newblock


\bibitem[Zeng et~al\mbox{.}(2019)]%
        {DBLP:conf/iccv/ZengHGTRZH19}
\bibfield{author}{\bibinfo{person}{Runhao Zeng}, \bibinfo{person}{Wenbing Huang}, \bibinfo{person}{Chuang Gan}, \bibinfo{person}{Mingkui Tan}, \bibinfo{person}{Yu Rong}, \bibinfo{person}{Peilin Zhao}, {and} \bibinfo{person}{Junzhou Huang}.} \bibinfo{year}{2019}\natexlab{}.
\newblock \showarticletitle{Graph Convolutional Networks for Temporal Action Localization}. In \bibinfo{booktitle}{\emph{{IEEE/CVF} International Conference on Computer Vision, {ICCV} 2019}}. \bibinfo{pages}{7093--7102}.
\newblock


\bibitem[Zhai et~al\mbox{.}(2020)]%
        {DBLP:conf/eccv/Zhai0TZY020}
\bibfield{author}{\bibinfo{person}{Yuanhao Zhai}, \bibinfo{person}{Le Wang}, \bibinfo{person}{Wei Tang}, \bibinfo{person}{Qilin Zhang}, \bibinfo{person}{Junsong Yuan}, {and} \bibinfo{person}{Gang Hua}.} \bibinfo{year}{2020}\natexlab{}.
\newblock \showarticletitle{Two-Stream Consensus Network for Weakly-Supervised Temporal Action Localization}. In \bibinfo{booktitle}{\emph{{ECCV} 2020 - 16th European Conference, Proceedings, Part {VI}}} \emph{(\bibinfo{series}{Lecture Notes in Computer Science}, Vol.~\bibinfo{volume}{12351})}. \bibinfo{pages}{37--54}.
\newblock


\bibitem[Zhang et~al\mbox{.}(2021b)]%
        {DBLP:conf/nips/ZhangWHWWOS21}
\bibfield{author}{\bibinfo{person}{Bowen Zhang}, \bibinfo{person}{Yidong Wang}, \bibinfo{person}{Wenxin Hou}, \bibinfo{person}{Hao Wu}, \bibinfo{person}{Jindong Wang}, \bibinfo{person}{Manabu Okumura}, {and} \bibinfo{person}{Takahiro Shinozaki}.} \bibinfo{year}{2021}\natexlab{b}.
\newblock \showarticletitle{FlexMatch: Boosting Semi-Supervised Learning with Curriculum Pseudo Labeling}. In \bibinfo{booktitle}{\emph{Advances in Neural Information Processing Systems 34: Annual Conference on Neural Information Processing Systems 2021, NeurIPS 2021}}. \bibinfo{pages}{18408--18419}.
\newblock


\bibitem[Zhang et~al\mbox{.}(2021a)]%
        {zhang2021cola}
\bibfield{author}{\bibinfo{person}{Can Zhang}, \bibinfo{person}{Meng Cao}, \bibinfo{person}{Dongming Yang}, \bibinfo{person}{Jie Chen}, {and} \bibinfo{person}{Yuexian Zou}.} \bibinfo{year}{2021}\natexlab{a}.
\newblock \showarticletitle{CoLA: Weakly-Supervised Temporal Action Localization With Snippet Contrastive Learning}. In \bibinfo{booktitle}{\emph{{IEEE} Conference on Computer Vision and Pattern Recognition, {CVPR} 2021}}. \bibinfo{pages}{16010--16019}.
\newblock


\bibitem[Zhang et~al\mbox{.}(2022)]%
        {DBLP:conf/eccv/ZhangWL22}
\bibfield{author}{\bibinfo{person}{Chen{-}Lin Zhang}, \bibinfo{person}{Jianxin Wu}, {and} \bibinfo{person}{Yin Li}.} \bibinfo{year}{2022}\natexlab{}.
\newblock \showarticletitle{ActionFormer: Localizing Moments of Actions with Transformers}. In \bibinfo{booktitle}{\emph{{ECCV} 2022 - 17th European Conference, Proceedings, Part {IV}}} \emph{(\bibinfo{series}{Lecture Notes in Computer Science}, Vol.~\bibinfo{volume}{13664})}. \bibinfo{pages}{492--510}.
\newblock


\bibitem[Zhao et~al\mbox{.}(2017)]%
        {DBLP:conf/iccv/ZhaoXWWTL17}
\bibfield{author}{\bibinfo{person}{Yue Zhao}, \bibinfo{person}{Yuanjun Xiong}, \bibinfo{person}{Limin Wang}, \bibinfo{person}{Zhirong Wu}, \bibinfo{person}{Xiaoou Tang}, {and} \bibinfo{person}{Dahua Lin}.} \bibinfo{year}{2017}\natexlab{}.
\newblock \showarticletitle{Temporal Action Detection with Structured Segment Networks}. In \bibinfo{booktitle}{\emph{{IEEE} International Conference on Computer Vision, {ICCV} 2017}}. \bibinfo{pages}{2933--2942}.
\newblock


\bibitem[Zhao et~al\mbox{.}(2020)]%
        {DBLP:journals/ijcv/ZhaoXWWTL20}
\bibfield{author}{\bibinfo{person}{Yue Zhao}, \bibinfo{person}{Yuanjun Xiong}, \bibinfo{person}{Limin Wang}, \bibinfo{person}{Zhirong Wu}, \bibinfo{person}{Xiaoou Tang}, {and} \bibinfo{person}{Dahua Lin}.} \bibinfo{year}{2020}\natexlab{}.
\newblock \showarticletitle{Temporal Action Detection with Structured Segment Networks}.
\newblock \bibinfo{journal}{\emph{Int. J. Comput. Vis.}} \bibinfo{volume}{128}, \bibinfo{number}{1} (\bibinfo{year}{2020}), \bibinfo{pages}{74--95}.
\newblock


\bibitem[Zheng et~al\mbox{.}(2020)]%
        {DBLP:conf/aaai/ZhengWLLYR20}
\bibfield{author}{\bibinfo{person}{Zhaohui Zheng}, \bibinfo{person}{Ping Wang}, \bibinfo{person}{Wei Liu}, \bibinfo{person}{Jinze Li}, \bibinfo{person}{Rongguang Ye}, {and} \bibinfo{person}{Dongwei Ren}.} \bibinfo{year}{2020}\natexlab{}.
\newblock \showarticletitle{Distance-IoU Loss: Faster and Better Learning for Bounding Box Regression}. In \bibinfo{booktitle}{\emph{The Thirty-Fourth {AAAI} Conference on Artificial Intelligence, {AAAI} 2020}}. \bibinfo{publisher}{{AAAI} Press}, \bibinfo{pages}{12993--13000}.
\newblock


\bibitem[Zhou et~al\mbox{.}(2023)]%
        {DBLP:conf/cvpr/ZhouHWLL23}
\bibfield{author}{\bibinfo{person}{Jingqiu Zhou}, \bibinfo{person}{Linjiang Huang}, \bibinfo{person}{Liang Wang}, \bibinfo{person}{Si Liu}, {and} \bibinfo{person}{Hongsheng Li}.} \bibinfo{year}{2023}\natexlab{}.
\newblock \showarticletitle{Improving Weakly Supervised Temporal Action Localization by Bridging Train-Test Gap in Pseudo Labels}. In \bibinfo{booktitle}{\emph{{IEEE/CVF} Conference on Computer Vision and Pattern Recognition, {CVPR} 2023}}. \bibinfo{pages}{23003--23012}.
\newblock


\bibitem[Zhu et~al\mbox{.}(2021)]%
        {DBLP:conf/iccv/ZhuT00021}
\bibfield{author}{\bibinfo{person}{Zixin Zhu}, \bibinfo{person}{Wei Tang}, \bibinfo{person}{Le Wang}, \bibinfo{person}{Nanning Zheng}, {and} \bibinfo{person}{Gang Hua}.} \bibinfo{year}{2021}\natexlab{}.
\newblock \showarticletitle{Enriching Local and Global Contexts for Temporal Action Localization}. In \bibinfo{booktitle}{\emph{{IEEE/CVF} International Conference on Computer Vision, {ICCV} 2021}}. \bibinfo{pages}{13496--13505}.
\newblock


\end{thebibliography}
